\documentclass[10pt,twocolumn,letterpaper]{article}

\usepackage{cvpr}      

\usepackage{xcolor}
\usepackage{graphicx} 
\usepackage{pifont}

\newcommand{\ra}[1]{\renewcommand{\arraystretch}{#1}}
\usepackage{adjustbox}
\usepackage{bibentry}

\usepackage{pifont}
\newcommand{\xmark}{\ding{55}}%

\usepackage{multirow} 

\usepackage{colortbl}
\usepackage{pgfplotstable}
\usepackage{siunitx}  

\usepackage{makecell}
\usepackage{tabularx}

\usepackage{pifont} 
\usepackage{amsmath} 

\definecolor{highlightblue}{RGB}{235, 240, 250} 
\definecolor{highlightgray}{gray}{0.92}        

\usepackage[ruled,vlined,linesnumbered]{algorithm2e}
\usepackage{amsmath}
\usepackage{xcolor}

\usepackage{newfloat}
\usepackage{listings}
\usepackage{makecell}
\usepackage{amsmath}
\DeclareCaptionStyle{ruled}{labelfont=normalfont,labelsep=colon,strut=off} 
\AtBeginDocument{
  
}

\lstdefinestyle{mystyle}{frame=single}    
\usepackage{subcaption}

\definecolor{cvprblue}{rgb}{0.21,0.49,0.74}
\usepackage[pagebackref,breaklinks,colorlinks,allcolors=cvprblue]{hyperref}

\usepackage{orcidlink}

\newcommand{\model}{\emph{{STAMP}}\xspace}

\title{Better, Stronger, Faster: Tackling the Trilemma in MLLM-based Segmentation with {Simultaneous Textual Mask Prediction}}
\vspace{-0.4em}
\author{
    Jiazhen Liu\orcidlink{0000-0003-0584-4571} \quad 
    Mingkuan Feng\thanks{Work done during his visit at HKUST.} \orcidlink{0009-0000-9282-0695} \quad 
    Long Chen\thanks{Corresponding author.} \orcidlink{0000-0001-6148-9709} \\[0.3ex]
    The Hong Kong University of Science and Technology \\[0.5ex]
    \tt\small jliugj@connect.ust.hk, fmk24@mails.tsinghua.edu.cn, longchen@ust.hk \\
    \small \url{https://github.com/HKUST-LongGroup/STAMP}
}

\begin{document}
\maketitle
\begin{abstract}

Integrating segmentation into Multimodal Large Language Models (MLLMs) presents a core trilemma: simultaneously preserving dialogue ability, achieving high segmentation performance, and ensuring fast inference. Prevailing paradigms are forced into a compromise. Embedding prediction methods introduce a conflicting pixel-level objective that degrades the MLLM's general dialogue abilities. The alternative, next-token prediction, reframes segmentation as an autoregressive task, which preserves dialogue but forces a trade-off between poor segmentation performance with sparse outputs or prohibitive inference speeds with rich ones. We resolve this trilemma with \textbf{all-mask prediction}, a novel paradigm that decouples autoregressive dialogue generation from non-autoregressive mask prediction. We present \model{\footnote{\model: \textbf{S}imultaneous \textbf{T}extual \textbf{A}ll-\textbf{M}ask \textbf{P}rediction. \label{footnote:model}}}, an MLLM that embodies this paradigm. After generating a textual response, \model predicts an entire segmentation mask in a single forward pass by treating it as a parallel “fill-in-the-blank" task over image patches. This design maintains the MLLM's dialogue ability by avoiding conflicting objectives, enables high segmentation performance by leveraging rich, bidirectional spatial context for all mask tokens, and achieves exceptional speed. Extensive experiments show that \model significantly outperforms state-of-the-art methods across multiple segmentation benchmarks, providing a solution that excels in dialogue, segmentation, and speed without compromise.

\end{abstract}    
\section{Introduction}
\label{sec:intro}

\begin{figure}[t]
    \centering
    \includegraphics[width=0.75\linewidth]{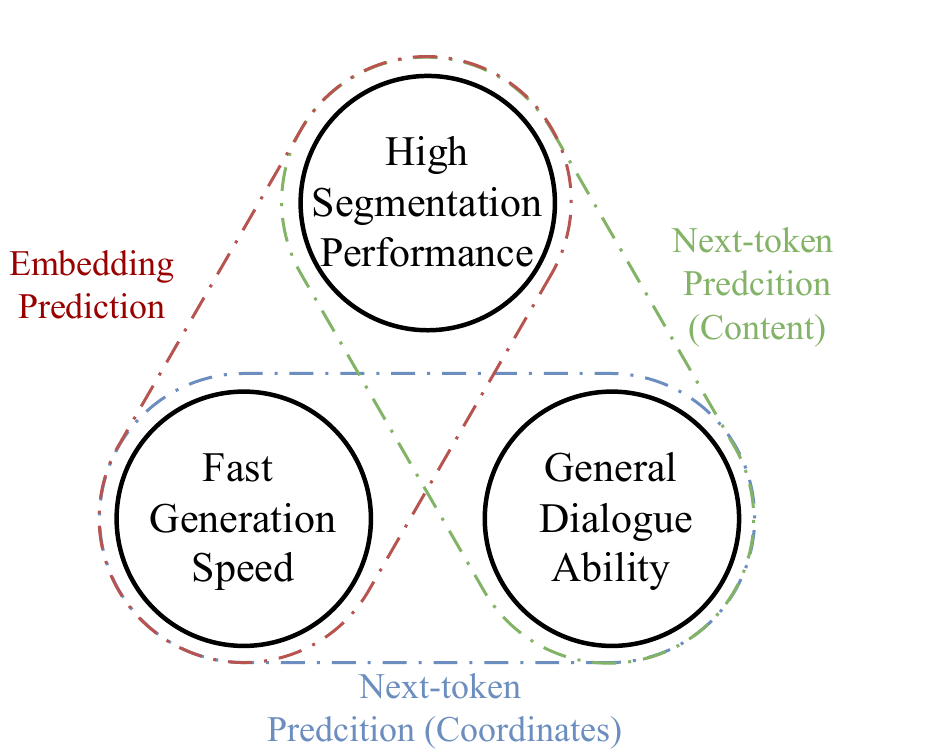}
    \caption{\textbf{The trilemma of segmentation in MLLMs.} Embedding prediction may harm dialogue abilities. Next-token prediction methods are either fast with poor segmentation performance or achieve superior performance at the cost of inference speed, particularly when generating rich content (\eg, chain-of-thought or patch-wise classification).}
    \label{fig:tri_problem}
    \vspace{-0.5em}
\end{figure}

\begin{table}[t]
    \centering
    \renewcommand{\arraystretch}{1} 
    \caption{\textbf{Comparison of MLLM segmentation paradigms.} “Steps" refers to mask generation steps (excluding textual prefix). “Decoder-free" indicates whether an external mask decoder is necessary. “Patch Class." refers to patch-wise classification.}
    
    \begin{adjustbox}{width=\linewidth}
    \setlength{\tabcolsep}{4pt} 
    \begin{tabular}{@{}llrrr@{}}
        \toprule
        \textbf{Model} & \textbf{\makecell[l]{Output}} & \textbf{\makecell[r]{Steps}} & \textbf{\makecell[r]{Token-only\\Supv.}} & \textbf{\makecell[r]{Decoder-\\free}} \\ 
        \midrule
        
        \multicolumn{5}{l}{\textit{\textbf{Paradigm 1: Embedding Prediction}}} \\
        LISA \small{(CVPR'24)}~\cite{lai2024lisa} & Embeddings & $\mathcal{O}(1)$ & \xmark & \xmark \\
        GSVA \small{(CVPR'24)}~\cite{xia2024gsva} & Embeddings & $\mathcal{O}(1)$ & \xmark & \xmark \\
        PixelLM \small{(CVPR'24)}~\cite{ren2024pixellm} & Embeddings & $\mathcal{O}(1)$ & \xmark & \xmark \\
        M$^2$SA \small{(ICLR'25)}~\cite{jang2025mmr} & Embeddings & $\mathcal{O}(1)$ & \xmark & \xmark \\
        READ \small{(CVPR'25)}~\cite{read} & Embeddings & $\mathcal{O}(1)$ & \xmark & \xmark \\
        \midrule

        \multicolumn{5}{l}{\textit{\textbf{Paradigm 2: Next-token Prediction}}} \\
        VisionLLM \small{(NIPS'24)}~\cite{wu2024visionllm} & Coordinates & $\mathcal{O}(N_{\text{points}})$ & \checkmark & \checkmark \\
        Seg-Zero \small{(arXiv'25)}~\cite{liu2025seg} & CoT + Coords. & $\mathcal{O}(N_{\text{CoT}})$ & \xmark & \xmark \\
        SegAgent \small{(CVPR'25)}~\cite{zhu2025segagent} & CoT + Coords. & $\mathcal{O}(N_{\text{CoT}})$ & \checkmark & \xmark \\
        Text4Seg \small{(ICLR'25)}~\cite{lan2024text4seg} & Patch Class. & $\mathcal{O}(N_{\text{patches}})$ & \checkmark & \checkmark \\
        \midrule

        \multicolumn{5}{l}{\textit{\textbf{Paradigm 3: All-mask Prediction (Ours)}}} \\
        \rowcolor{highlightgray}
        \model (Ours) & Patch Class. & $\mathcal{O}(1)$ & \checkmark & \checkmark \\

        \bottomrule
    \end{tabular}
    \end{adjustbox}
    \label{tab:comp-paradigm}
\end{table}

The success of Multimodal Large Language Models (MLLMs) has spurred a trend to unify diverse vision tasks within a single instruction-driven framework~\cite{yang2025qwen3, chen2024lion, zhang2025mllms, jiang2025detect, zhu2025internvl3, li2024llava}. For such models to be practical, they must resolve a core trilemma: simultaneously (i) preserving dialogue abilities, (ii) achieving high task performance, and (iii) ensuring fast inference. While this has been successful for recognition~\cite{zhu2025internvl3, li2024llava} and detection~\cite{bai2025qwen2, yang2025qwen3}, the trilemma remains unsolved for segmentation. This stems from a fundamental challenge: the MLLM's sequential text-generative nature is ill-suited for producing dense, pixel-level masks~\cite{liu2025segmentation, lan2024text4seg}. Consequently, current MLLM segmentation paradigms are forced into a trade-off, where they must compromise on one or more of these fronts.

\begin{figure*}[t]
    \centering
    \includegraphics[width=\linewidth]{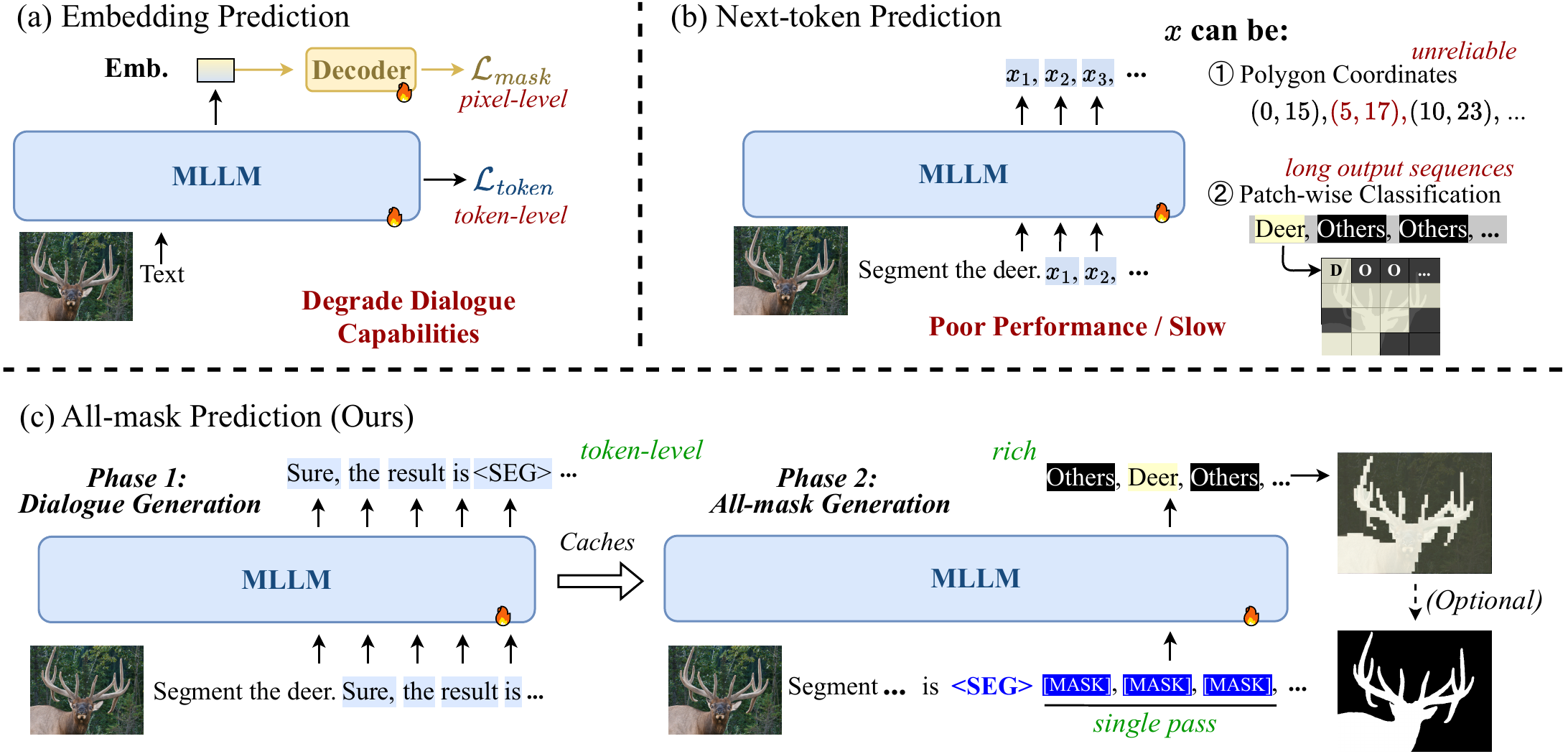}
\caption{\textbf{Comparison of MLLM-based segmentation paradigms.} 
\textbf{(a) Embedding Prediction}: A conflicting pixel-level objective~\cite{lai2024lisa, ren2024pixellm} degrades the MLLM's general dialogue capabilities.
\textbf{(b) Next-token Prediction}: Generates masks autoregressively~\cite{wang2023visionllm, liu2025seg, lan2024text4seg}, forcing a trade-off between poor segmentation performance (for sparse outputs) and slow inference (for rich outputs).
\textbf{(c) Our All-mask Prediction}: We decouple dialogue generation (autoregressive) from mask generation (non-autoregressive). By simultaneously predicting all mask tokens as patch-wise classifications in a single pass, our paradigm resolves the segmentation trilemma, uniting preserved dialogue abilities, high segmentation performance and fast inference speed.
}
    \label{fig:intro}
\end{figure*}

\cref{fig:tri_problem} illustrates how mainstream MLLM segmentation paradigms tackle this trilemma, with their primary differences rooted in the MLLM's role and output format (\cf \cref{tab:comp-paradigm}). The first paradigm, \emph{embedding prediction} (\cf \cref{fig:intro}a), achieves good segmentation performance by fine-tuning the MLLM with a pixel-level mask loss to control an external mask decoder~\citep{lai2024lisa, ren2024pixellm, read, xia2024gsva, rasheed2024glamm}. However, this auxiliary objective degrades general dialogue abilities~\citep{lan2024text4seg, liu2025segmentation, wu2024see}, as seen in models like LISA~\citep{lai2024lisa}, which may fail to follow a simple instruction like ``How many deer are there?" and instead output a segmentation result~\cite{liu2025segmentation}.

The second paradigm, next-token prediction (\cf \cref{fig:intro}b), avoids this objective conflict by reframing segmentation as a pure language modeling task, where the MLLM autoregressively generates a textual representation of the mask. An early approach was to generate sparse \emph{token-to-coordinates} outputs, such as polygon vertices defining an object's contour~\citep{wang2023visionllm}. This method, however, proved unreliable due to its high susceptibility to error accumulation, where a single inaccurate point can corrupt the entire mask~\cite{wu2024visionllm, lan2024text4seg}, leading to poor segmentation performance. To mitigate this unreliability, a recent \emph{token-to-content} trend generates richer outputs, like Chain-of-Thought (CoT) reasoning~\cite{liu2025seg, zhu2025segagent} or patch-wise foreground/background classifications (\cf \cref{fig:intro}b)~\cite{lan2024text4seg}. While these strategies improve segmentation performance, the autoregressive generation of extremely long token sequences results in prohibitively slow inference speeds.

Evidently, existing paradigms face an inherent trade-off among dialogue ability, segmentation performance, and inference speed. In this work, we break this trilemma by formulating mask generation as a non-autoregressive, patch-wise classification task. First, this approach keeps supervision at the token level by defining each mask token as a textual classification for a corresponding image patch, thus preserving the MLLM's dialogue capabilities. Second, the direct one-to-one correspondence between mask tokens and image patches provides the rich representation required for high segmentation performance. Finally, by fixing the length of mask tokens to match the number of image patches, we reframe mask generation as a fill-in-the-blank task. This enables the simultaneous prediction of all mask tokens over pre-defined placeholders in a single forward pass, achieving exceptional inference speed.

Thus, we introduce \textbf{all-mask prediction} (\cf \cref{fig:intro}c), a novel paradigm that decouples autoregressive dialogue from non-autoregressive mask generation. Beyond resolving the trilemma, this paradigm offers two critical advantages. For one, by generating all mask tokens in a single pass, our method ensures that every token is mutually visible during prediction. This allows the model to leverage a rich, bidirectional 2D spatial context, overcoming the unidirectional limitation inherent in autoregressive approaches. Moreover, by defining placeholders (\texttt{[MASK]}) based on the input image, our method inherits a natural compatibility with dynamic resolutions~\citep{bai2025qwen2, yang2025qwen3, zhu2025internvl3}.

To validate our paradigm, we implement \model{\footref{footnote:model}} model. After generating a textual response, the model can emit a special \texttt{<SEG>} token. This triggers a single-pass prediction stage where bidirectional attention is applied over all mask placeholders, whose initial embeddings are fused with their corresponding image patch features. The effectiveness of this approach is immediate: on the standard referring segmentation benchmark, \model{} establishes a new state-of-the-art, outperforming the strongest prior methods by 4.5\%. Crucially, it achieves this with fast inference speeds on par with the embedding prediction methods, all while maintaining exceptional dialogue ability.

In summary, our main contributions are threefold:
\begin{itemize}
\item We introduce \emph{all-mask prediction}, a novel prediction paradigm that resolves the segmentation trilemma by avoiding the trade-offs of prior autoregressive and embedding-based methods.
\item We propose \model, the first realization of this paradigm, demonstrating how to seamlessly integrate autoregressive dialogue with non-autoregressive mask generation.
\item \model achieves state-of-the-art results across multiple segmentation benchmarks, significantly advancing performance and speed while preserving the MLLM's core dialogue capabilities.
\end{itemize}

\section{Related Work}
\label{sec:rel}

\noindent\textbf{Multimodal Large Language Models (MLLMs).} The advent of MLLMs, powered by the advanced reasoning of their LLM foundations~\citep{kaplan2020scaling, openai2024hello, gemini}, has established a new frontier for instruction-driven vision tasks, including dense prediction like segmentation. Architectures such as LLaVA~\citep{llava, liu2024improved, li2024llava}, InstructBLIP~\citep{dai2023instructblip}, and Qwen-VL~\citep{bai2023qwenvl, bai2025qwen2, yang2025qwen3} typically bridge a visual encoder with a pre-trained LLM, enabling them to ground complex language instructions within an image's spatial context. The central challenge for segmentation, therefore, is how to translate this high-level, spatially-aware understanding into a pixel-level mask. Two primary strategies have emerged to address this: one leverages the MLLM as a powerful vision-language encoder, using its internal embeddings to guide a separate mask decoder~\citep{lai2024lisa, xia2024gsva}; the other reformulates mask generation as a language task, compelling the MLLM to ``describe" the mask through text~\citep{lan2024text4seg, wu2024visionllm, liu2025seg}.

\noindent\textbf{MLLM-based Segmentation.} Existing MLLM-based segmentation methods can be categorized into two primary paradigms: \emph{Embedding Prediction} and \emph{Next-token Prediction} (\cf \cref{tab:comp-paradigm}). They are distinguished by the MLLM's output format for mask generation. The former employs the MLLM to produce continuous embeddings that guide an external mask decoder. The latter, however, uses the MLLM to generate a sequence of discrete tokens which, in turn, define the segmentation mask.

\noindent\underline{\emph{Embedding Prediction.}} This paradigm, pioneered by LISA\citep{lai2024lisa}, trains an MLLM to output a special token (\eg, \texttt{[SEG]}) whose embedding prompts an external, SAM-like decoder~\citep{ravi2024sam}. As noted in \cref{tab:comp-paradigm}, this approach, refined by subsequent works~\cite{ren2024pixellm, xia2024gsva, jang2025mmr, read, wang2025segllm}, yields an efficient $\mathcal{O}(1)$ mask generation step, as the MLLM only needs to generate a single special token to initiate the entire process. However, this design introduces significant drawbacks. The mandatory external module means these methods are not mask decoder-free, requiring structural modifications to the MLLM architecture. Furthermore, training this decoder violates the principle of token-only supervision by necessitating a pixel-level loss, which often compromises the MLLM's general dialogue abilities.

\noindent\underline{\emph{Next-token Prediction.}} This paradigm avoids objective conflicts by representing the mask as a sequence of discrete tokens. Early works like VisionLLM~\citep{wang2023visionllm} generated sparse polygon coordinates, a natively decoder-free design. However, this approach proved unreliable due to error accumulation. To improve robustness, methods like Seg-Zero~\citep{liu2025seg} and SegAgent~\citep{zhu2025segagent} use CoT reasoning to plan keypoints before calling an external SAM decoder. A different strategy, proposed by Text4Seg~\citep{lan2024text4seg}, generates patch-wise classifications to remain decoder-free. Despite these advances, the entire paradigm is fundamentally bottlenecked by its autoregressive nature. As shown in \cref{tab:comp-paradigm}, the number of generation steps scales linearly with the sequence length, making it prohibitively slow for the long outputs required for high-quality segmentation.

\section{Method}

\begin{figure*}[t]
    \centering
    \includegraphics[width=0.9\linewidth]{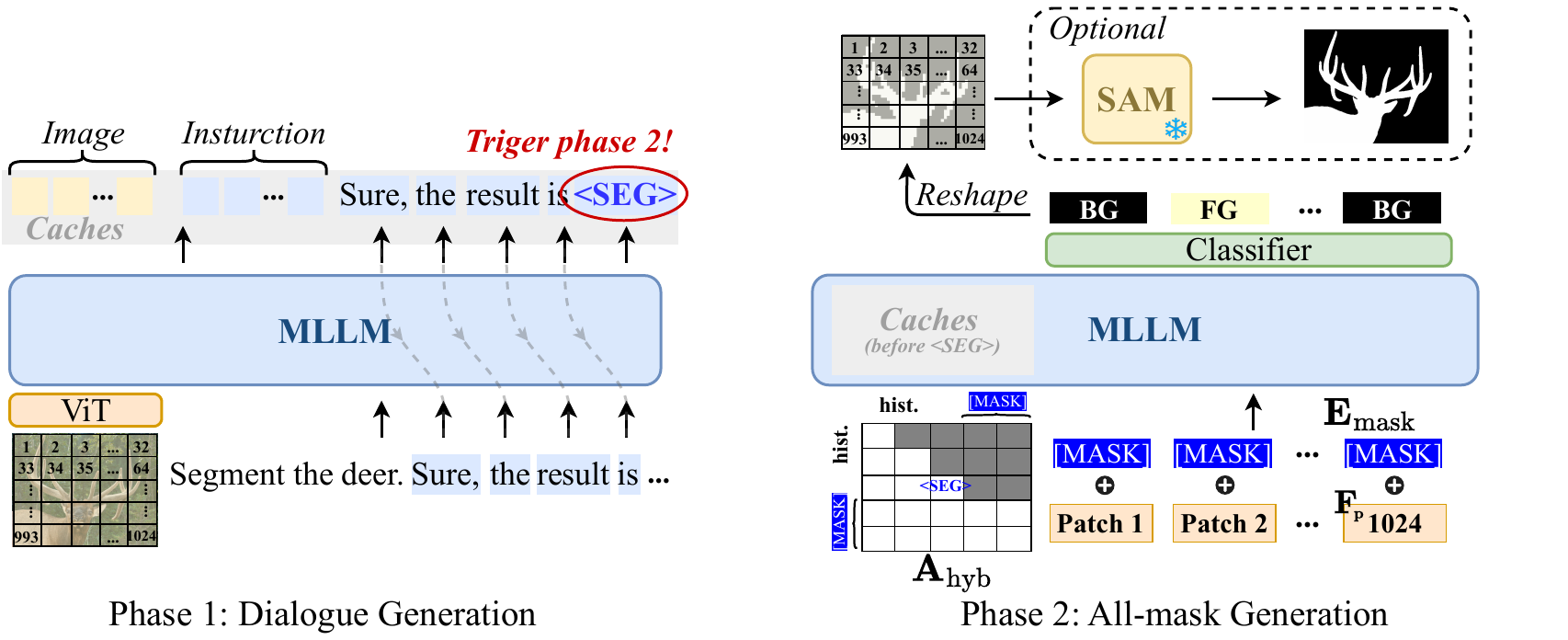}
    \caption{\textbf{The \model{} Pipeline.} 
\textbf{Phase 1 (Dialogue Generation):} The MLLM autoregressively generates a conversational response, emitting a special \texttt{<SEG>} token to trigger mask generation. 
\textbf{Phase 2 (All-mask Generation):} Triggered by the \texttt{<SEG>} token, a sequence of \texttt{[MASK]} placeholders, each fused with its corresponding image patch feature, is processed. A single non-autoregressive forward pass with hybrid attention then predicts all mask tokens simultaneously.
}
    \label{fig:methods}
\end{figure*}

In this section, we introduce \model{}, our framework for realizing the \emph{All-mask Prediction} paradigm. The key challenge is determining \emph{when} to generate the mask and \emph{where} to place its tokens. As depicted in \cref{fig:methods}, our solution is a two-phase architecture. In Phase 1 (Dialogue Generation), the MLLM operates autoregressively to generate a textual response, learning to emit a special \texttt{<SEG>} token. This single token elegantly resolves the core challenge: its presence signals \emph{when} to segment, while its position dictates \emph{where} the mask placeholders will be inserted. The emission of \texttt{<SEG>} triggers Phase 2 (All-mask Generation), which is highly efficient as it leverages the cached activations from the initial phase. In a single, non-autoregressive forward pass, it predicts a classification (\eg, foreground/background) for each image patch based on a sequence of \texttt{[MASK]} placeholders fused with their respective patch features. The complete pipeline is detailed in Algorithm~\ref{alg:pipeline_a2e}, and we elaborate on each phase below.

\newcommand{\vect}[1]{\mathbf{#1}}
\newcommand{\set}[1]{\mathcal{#1}}
\newcommand{\code}[1]{\texttt{#1}}

\subsection{Phase 1: Dialogue Generation}
\label{sec:phase1}

In Phase 1, the model generates a contextual textual response and identifies segmentation targets. Given an image $I$ and an instruction $T$, a ViT first extracts patch features $\mathbf{F_p} \in \mathbb{R}^{N \times D}$, which are then prepended to the text embeddings. The MLLM processes this combined input to autoregressively generate a response $R$. During this process, the model can emit a special \texttt{<SEG>} token from its vocabulary. Critically, unlike methods such as LISA~\citep{lai2024lisa} that tie the \texttt{[SEG]} token's embedding to an external decoder, our \texttt{<SEG>} token is a standard vocabulary item. It functions purely as a learned, in-vocabulary signal that triggers Phase 2, as depicted in~\cref{fig:methods} (left).

\textbf{Caches $\set{C}$.} To ensure an efficient transition, intermediate representations from Phase 1 are cached. For each \texttt{<SEG>} token, we define a context-specific tuple $(\text{hist}_i, \text{cache}_i)$, where $\text{hist}_i$ is the dialogue history leading to the token, and $\text{cache}_i$ stores the pre-computed key-value (KV) states for all preceding tokens. Preserving these states allows \model{} to avoid redundant computation when initiating mask prediction for each target.

\begin{algorithm}[htbp]
\SetKwComment{Comment}{$\triangleright$\ }{}
\SetCommentSty{color=gray}
\SetKwInOut{Require}{Require}
\SetKwInOut{Ensure}{Ensure}
\DontPrintSemicolon

\caption{\model{} Inference Pipeline}
\label{alg:pipeline_a2e}

\Require{Image $I$, Instruction $T$}
\Ensure{Textual response $R$, Set of masks $\set{M}$}

\BlankLine
\textbf{Phase 1: Dialogue Generation}\;
$\vect{F_p} \gets \text{ViT}(I)$\;
$R, \set{C} \gets \text{GenerateAutoregressive}(T, \vect{F_p})$ \Comment*{Generate response and capture caches}

\BlankLine
\textbf{Phase 2: All-mask Generation}\;
$\set{M} \gets \emptyset$\;
\For{each $(\text{hist}_i, \text{cache}_i)$ in $\set{C}$}{
    $\vect{E}_{\text{mask}} \gets \text{Embed}(\code{[MASK]}_{1..N}) + \vect{F_p}$ \Comment*{Create visually-augmented mask tokens}
    $\vect{S}_{\text{in}} \gets \text{Concat}(\text{Embed}(\text{hist}_i), \vect{E}_{\text{mask}})$\;
    $\vect{A}_{\text{hyb}} \gets \text{ConstructHybridMask}(\text{len}(\text{hist}_i), N)$\;
    $\vect{Z}_{\text{mask}} \gets \text{MLLM.forward}(\mathbf{S}_{\text{in}}, \text{cache}=\text{cache}_i,$ \;
    \qquad $\text{attn\_mask}=\mathbf{A}_{\text{hyb}})$
    
    $\text{logits} \gets \text{Classifier}(\vect{Z}_{\text{mask}})$\;
    $\text{mask}_i \gets \text{argmax}(\text{logits})$\;
    \textcolor{gray}{$\text{mask}_i \gets \text{Refine}(\text{mask}_i, \text{SAM})$} \Comment*{{Optional}}
    $\set{M} \gets \set{M} \cup \{\text{mask}_i\}$\;
}
\Return{$R, \set{M}$}
\end{algorithm}

\subsection{Phase 2: All-mask Generation}
\label{sec:phase2}

Upon the emission of a \texttt{<SEG>} token, Phase 2 generates the entire mask in a single, non-autoregressive forward pass. This begins by preparing a specialized input sequence $\mathbf{S}_\text{in}$. We take the dialogue history preceding the \texttt{<SEG>} token and append $N$ \texttt{[MASK]} placeholders, one for each image patch. To provide spatial relationship, the initial embedding of each \texttt{[MASK]} token is fused with its corresponding ViT patch feature from $\mathbf{F_p}$, along with the positional embedding that specifies the patch's location in the original image grid. We term these visually-augmented mask embeddings $\mathbf{E}_{\text{mask}}$.

A key component of this phase is our hybrid attention mechanism, which uses a custom attention mask $\mathbf{A}_\text{hyb}$ to partition the sequence. For the dialogue history, it enforces standard causal attention. For the $\mathbf{E}_{\text{mask}}$, it enables bi-directional attention, allowing each placeholder to attend to the entire context and all other placeholders. This ensures a holistic and context-aware prediction, as shown in~\cref{fig:methods}.

With this setup, the MLLM performs a single forward pass. The KV states for the dialogue history are efficiently loaded from the Phase 1 cache. The final hidden states $Z_\text{mask}$ corresponding to the \texttt{[MASK]} input are then fed into a linear classifier to produce foreground (FG) or background (BG) logits for each patch. Finally, this patch-level output can be optionally refined into a high-resolution mask by sampling a single keypoint from the prediction to prompt a frozen SAM decoder~\citep{lan2024text4seg}.

\subsection{Training}
We train \model{} end-to-end with a unified token-level prediction objective comprising two components: a text generation loss ($\mathcal{L}_{\text{text}}$) and a mask prediction loss ($\mathcal{L}_{\text{mask}}$).

The text loss $\mathcal{L}_{\text{text}}$ is the standard cross-entropy loss for autoregressive language modeling. For a ground-truth response $Y = (y_1, \dots, y_L)$, it maximizes the log-likelihood of each token:
\begin{equation}
    \mathcal{L}_{\text{text}} = - \sum_{i=1}^{L} \log P(y_i | y_{<i}, I, T).
\end{equation}

We extend this token-level supervision to the mask generation phase. The mask loss $\mathcal{L}_{\text{mask}}$ is applied to the output logits of the $\mathbf{E}_{\text{mask}}$. Each token performs a binary classification on its corresponding image patch, predicting whether it belongs to the foreground or background. This is supervised using a combination of Binary Cross-Entropy and Dice loss:
\begin{equation}
    \mathcal{L}_{\text{mask}} = \mathcal{L}_{\text{BCE}} + \mathcal{L}_{\text{Dice}}.
\end{equation}
    
The final training objective is a simple sum of the two losses: $\mathcal{L} = \mathcal{L}_{\text{text}} + \mathcal{L}_{\text{mask}}$. By jointly optimizing both components, \model{} learns to treat segmentation not as a separate task, but as an integral part of its generative vocabulary.


\begin{table*}[h]
    \centering
    \renewcommand{\arraystretch}{0.95} 
    \small
    \caption{\textbf{Comprehensive comparison on the single-object RES task.} Methods are categorized by paradigm and sorted by performance. Our models are highlighted with a gray background, and the top-performing method is shown in bold. “Mask-Decoder-Free" indicates results without SAM-based post-processing.}
    \label{tab:final_with_citations_sorted}
    
    \definecolor{highlightgray}{gray}{0.93}
    \definecolor{citegray}{gray}{0.55}
    \newcommand{\citeinfo}[1]{{\color{citegray}\scriptsize #1}}

    \begin{adjustbox}{max width=0.82\linewidth} 
    \setlength{\tabcolsep}{6pt}
    \begin{tabular}{@{}llrrrrrrrrr@{}}
        \toprule
        \multirow{2}{*}{\textbf{Method}} & \multirow{2}{*}{\textbf{LLM}} & \multicolumn{3}{c}{\textbf{RefCOCO}} & \multicolumn{3}{c}{\textbf{RefCOCO+}} & \multicolumn{2}{c}{\textbf{RefCOCOg}} & \multirow{2}{*}{\textbf{Avg.}} \\
        \cmidrule(lr){3-5} \cmidrule(lr){6-8} \cmidrule(lr){9-10}
        & & val & testA & testB & val & testA & testB & val(U) & test(U) & \\
        \midrule

        \multicolumn{11}{@{}l}{\textbf{\textit{Specialised Baselines}}} \\
        \midrule
        HIPIE \citeinfo{(NIPS'24)~\cite{wang2023hierarchical}} & BERT & 78.3 & - & - & 66.2 & - & - & 69.8 & - & - \\
        ReLA \citeinfo{(CVPR'23)~\cite{liu2023gres}} & BERT & 73.8 & 76.5 & 70.2 & 66.0 & 71.0 & 57.7 & 65.0 & 66.0 & 68.3 \\
        PolyFormer-L \citeinfo{(CVPR'23)~\cite{liu2023polyformer}} & BERT & 76.0 & 78.3 & 73.3 & 69.3 & 74.6 & 61.9 & 69.2 & 70.2 & 71.6 \\
        UNINEXT-L\citeinfo{(CVPR'24)~\cite{yan2023universal}} & {BERT} & {80.3} & {82.6} & {77.8} & {70.0} & {74.9} & {62.6} & {73.4} & {73.7} & {74.4} \\
        \addlinespace

        \multicolumn{11}{@{}l}{\textbf{\textit{Paradigm: Embedding Prediction}}} \\
        \midrule
        PixelLM \citeinfo{(CVPR'24)~\cite{ren2024pixellm}} & Vicuna-7B & 73.0 & 76.5 & 68.2 & 66.3 & 71.7 & 58.3 & 69.3 & 70.5 & 69.2 \\
        LISA \citeinfo{(CVPR'24)~\cite{lai2024lisa}} & Vicuna-7B & 74.9 & 79.1 & 72.3 & 65.1 & 70.8 & 58.1 & 67.9 & 70.6 & 69.9 \\
        LISA \citeinfo{(CVPR'24)~\cite{lai2024lisa}} & Vicuna-13B & 76.0 & 78.8 & 72.9 & 65.0 & 70.2 & 58.1 & 69.5 & 70.5 & 70.1 \\
        GSVA \citeinfo{(CVPR'24)~\cite{xia2024gsva}} & Vicuna-7B & 77.2 & 78.9 & 73.5 & 65.9 & 69.6 & 59.8 & 72.7 & 73.3 & 71.4 \\
        READ \citeinfo{(CVPR'25)~\cite{read}} & Vicuna-7B & 78.1 & 80.2 & 73.2 & 68.4 & 73.7 & 60.4 & 70.1 & 71.4 & 71.9 \\
        {GSVA} \citeinfo{(CVPR'24)~\cite{xia2024gsva}} & {Vicuna-13B} & {78.2} & {80.4} & {74.2} & {67.4} & {71.5} & {60.9} & {74.2} & {75.6} & {72.8} \\
        \addlinespace

        \multicolumn{11}{@{}l}{\textbf{\textit{Paradigm: Token Prediction}}} \\
        \midrule
        \multicolumn{11}{@{}l}{\quad \textit{Mask-Decoder-Free}} \\
        Text4Seg \citeinfo{(ICLR'25)~\cite{lan2024text4seg}} & Vicuna-13B & 74.1 & 76.4 & 72.4 & 68.5 & 72.8 & 63.6 & 69.1 & 70.1 & 70.9 \\
        Text4Seg \citeinfo{(ICLR'25)~\cite{lan2024text4seg}} & InternLM2.5-7B & 74.7 & 77.4 & 71.6 & 68.5 & 73.6 & 62.9 & 70.7 & 71.6 & 71.4 \\
        \rowcolor{highlightgray}
        \model & Qwen2-2B & 77.7 & {79.4} & 76.1 & 73.4 & 76.4 & 69.7 & 74.9 & 75.1 & 75.3 \\
        \rowcolor{highlightgray}
        \model & {Qwen2-7B} & \textbf{78.1} & \textbf{79.2} & \textbf{76.8} & \textbf{74.7} & \textbf{77.6} & \textbf{70.9} & \textbf{75.7} & \textbf{76.2} & \textbf{76.2} \\
        \addlinespace[0.3em]
        
        \multicolumn{11}{@{}l}{\quad \textit{With Mask Decoder}} \\
        Seg-Zero \citeinfo{(arXiv'25)~\cite{liu2025seg}} & Qwen2.5-3B & - & 79.3 & - & - & 73.7 & - & - & 71.5 & - \\
        Seg-Zero \citeinfo{(arXiv'25)~\cite{liu2025seg}} & Qwen2.5-7B & - & 80.3 & - & - & 76.2 & - & - & 72.6 & - \\
        SegLLM \citeinfo{(ICLR'25)~\cite{wang2025segllm}} & Vicuna-7B & 80.2 &  81.5 &  75.4 &  70.3 &  73.0 &  62.5 &  72.6 &  73.6 &  73.6 \\
        Text4Seg \citeinfo{(ICLR'25)~\cite{lan2024text4seg}} & Vicuna-7B & 79.3 & 81.9 & 76.2 & 72.1 & 77.6 & 66.1 & 72.1 & 73.9 & 74.9 \\
        Text4Seg \citeinfo{(ICLR'25)~\cite{lan2024text4seg}} & InternLM2.5-7B & 79.2 & 81.7 & 75.6 & 72.8 & 77.9 & 66.5 & 74.0 & 75.3 & 75.4 \\
        SegAgent \citeinfo{(CVPR'25)~\cite{zhu2025segagent}} & Qwen-7B & 79.7 & 81.4 & 76.6 & 72.5 & 75.8 & 66.9 & 75.1 & 75.2 & 75.4 \\
        Text4Seg \citeinfo{(ICLR'25)~\cite{lan2024text4seg}} & Vicuna-13B & 80.2 & 82.7 & 77.3 & 73.7 & 78.6 & 67.6 & 74.0 & 75.1 & 76.2 \\
        \rowcolor{highlightgray}
        \model & Qwen2-2B & 81.9 & 83.7 & 79.5 & 77.1 & 80.5 & 72.7 & 78.5 & 78.8 & 79.1 \\
        \rowcolor{highlightgray}
        \model & {Qwen2-7B} & \textbf{83.1} & \textbf{84.5} & \textbf{80.8} & \textbf{79.4} & \textbf{82.8} & \textbf{74.6} & \textbf{79.9} & \textbf{80.4} & \textbf{80.7} \\
        \bottomrule
    \end{tabular}
    \end{adjustbox}
\end{table*}

\section{Experiments}

\subsection{Common Setup}

\noindent\textbf{Implementation Details.} We built \model{} on the 2B and 7B variants of the Qwen2-VL series~\cite{bai2025qwen2}, which provides inherent support for dynamic input resolutions. For the optional mask refinement stage, we employed the huge (-H) variant of SAM. The number of \texttt{[MASK]} placeholders was set dynamically between 1024 and 1280 to correspond with the input image resolution.

\noindent\textbf{Training.} All models were trained on NVIDIA H800 GPUs. We used a maximum learning rate of 3e-5, which followed a linear decay schedule after a warm-up phase constituting 3\% of the total training steps. The weight decay was set to 0, and gradient norms were clipped at 1.0. Following common practice, the 2B models were fully fine-tuned, while the larger 7B models were trained using LoRA~\cite{hu2022lora} for parameter-efficient fine-tuning.

With these implementation and training details established, we now proceed to a comprehensive evaluation of \model{}, which starts from referring and reasoning segmentation tasks, extending to more general dialogue abilities, and concluding with a detailed analysis of inference efficiency and key design choices via ablation studies.

\subsection{Referring Expression Segmentation}

\noindent\textbf{Setup.} We first evaluated \model{} on the task of Referring Expression Segmentation (RES) to validate its core segmentation performance. Our evaluation covered both single-object RES, using the standard RefCOCO~\citep{kazemzadeh2014referitgame}, RefCOCO+~\citep{kazemzadeh2014referitgame}, and RefCOCOg~\citep{mao2016generation} datasets, as well as the more challenging multi-object and no-object scenarios presented by gRefCOCO~\cite{liu2023gres}.

To ensure a fair and direct comparison with prior work, we strictly adhered to the training data and schedule of Text4Seg~\cite{lan2024text4seg}. Initially, the model was trained for 3 epochs on a combined dataset of 800k samples, comprising the training splits of RefCLEF~\cite{kazemzadeh2014referitgame}, RefCOCO, RefCOCO+, and RefCOCOg. Following this, we continued training for an additional 2 epochs exclusively on the 419k samples from the gRefCOCO training split. All testing was conducted using the standard evaluation protocols~\citep{lai2024lisa, lan2024text4seg}.

\noindent\textbf{Baselines.} As shown in~\cref{tab:final_with_citations_sorted}, we compare \model{} against a comprehensive set of recent and influential works. These include state-of-the-art \emph{Specialised Baselines} (\eg, UNINEXT-L~\citep{yan2023universal}) for context, as well as top methods from the two dominant MLLM paradigms: \emph{Embedding Prediction} (\eg, GSVA~\citep{xia2024gsva}, READ~\citep{read}) and \emph{Token Prediction} (\eg, next-token prediction method Text4Seg~\citep{lan2024text4seg}). For the multi-/no-object task on gRefCOCO, we also include specially designed models like LAVT~\citep{yang2022lavt}.

\noindent\textbf{Results on Single-Object RES.}
As presented in Table~\ref{tab:final_with_citations_sorted}, our \model{}-7B model set a new state of the art, achieved an average cIoU of 80.7 and outperformed all competing methods across all evaluation splits. This marked a significant advance over the previous best, Text4Seg-13B (76.2 cIoU). Notably, even our lightweight \model{}-2B variant surpassed this prior art with 79.1 cIoU, highlighting the efficiency of our approach. The core strength of our architecture is further evidenced by an unrefined version of \model{}-7B, which matched the previous state of the art without any SAM-based post-processing. Crucially, these performance gains were attained even though competitors like Seg-Zero and Text4Seg leveraged more powerful LLMs (Qwen2.5-7B and InternLM2.5-7B vs. our Qwen2), confirming that our improvements originate from a superior architectural design rather than simply a stronger backbone.

\noindent\textbf{Results on Multi/No-Object RES.}
As presented in Table~\ref{tab:gref}, \model{} extended its superior performance to the more challenging gRefCOCO benchmark. Echoing our previous findings, our \model{}-7B model without SAM-based refinement already surpassed the strongest fully-equipped baseline, Text4Seg (71.8 vs. 71.5). Our best-performing model established a new state of the art with a score of 74.8, exceeding the strongest competitor by a significant margin of 3.3\%. This robust performance underscores the effectiveness and reliability of our architecture for complex segmentation challenges involving multiple or absent objects.

\begin{figure*}[thbp]
    \centering
    \includegraphics[width=\linewidth]{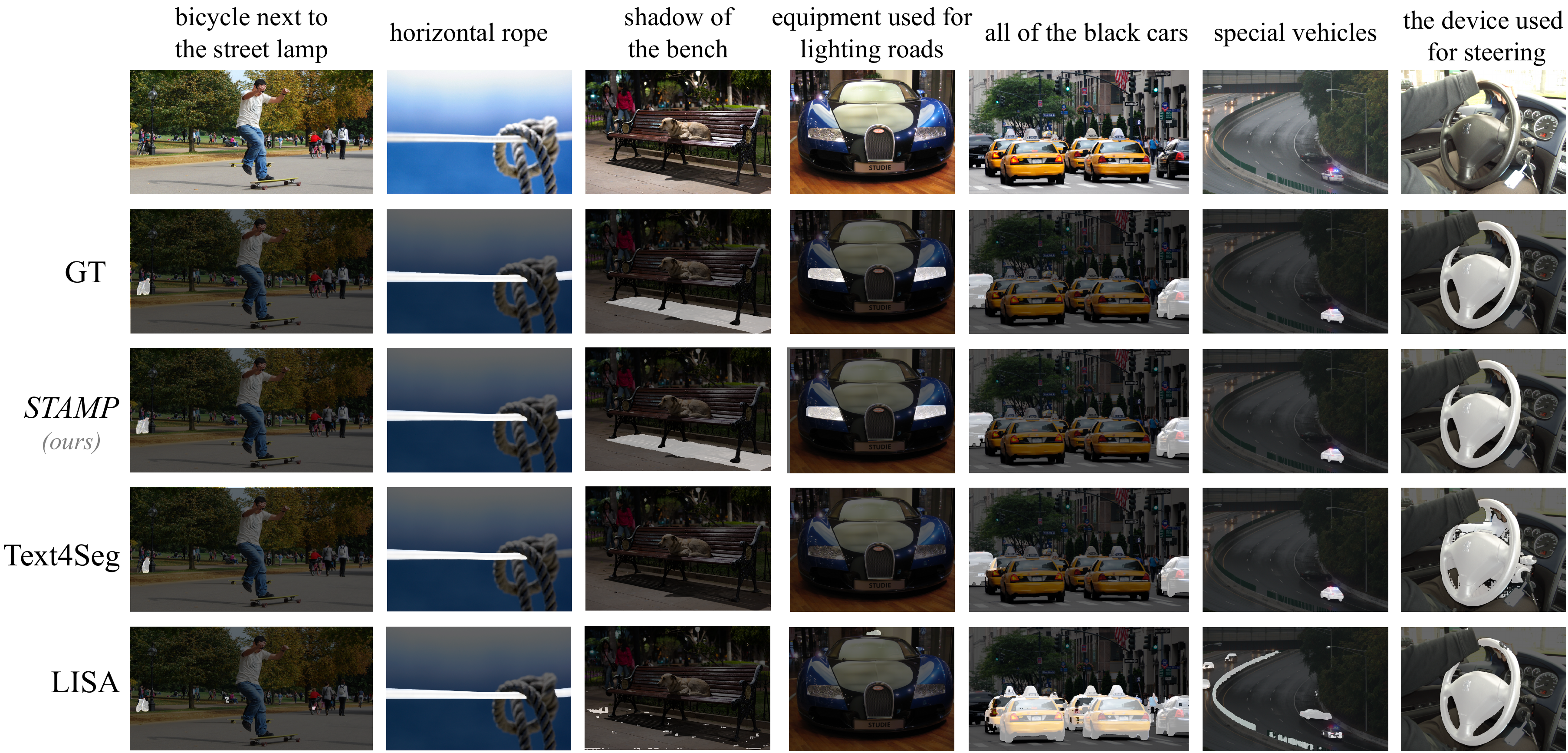}
    \caption{\textbf{Showcase of \model{} against competing methods.}}
    \label{fig:showcase}
\end{figure*}

\subsection{Reasoning Segmentation}

\noindent\textbf{Setup.} To evaluate the model's complex reasoning capabilities, we tested \model{} on the reasoning segmentation benchmark, following the protocol of LISA~\citep{lai2024lisa}. We further fine-tuned our single-object RES model (2B) for one epoch on a mixture of the RefCOCO* datasets and the 239 training images from the ReasonSeg training split~\citep{lai2024lisa}. No additional training data was used. Evaluation adhered to the established benchmark protocol from LISA. We include OVSeg~\cite{liang2023open} as a traditional baseline to highlight the difficulty of the ReasonSeg benchmark, as its challenges can typically only be addressed by MLLM-based models.

\noindent\textbf{Results.}
As shown in Table~\ref{tab:reasonseg}, our \model{}-2B model outperformed all existing methods, achieving an average score of 63.2. This result significantly surpassed the previous leading method, READ~\citep{read} (61.1), which utilized a larger training dataset~\citep{wu2024see}. Notably, this superior performance was achieved without explicit chain-of-thought reasoning and with a substantially smaller backbone (Qwen2-VL-2B) compared to competitors like Seg-Zero (Qwen2.5-VL-7B) and Text4Seg (Qwen2-VL-7B). This provides strong evidence that our all-mask prediction paradigm not only enhances segmentation but also more effectively leverages the MLLM's intrinsic reasoning capabilities. The qualitative results in ~\cref{fig:showcase} further substantiate this, showcasing accurate segmentation from complex instructions.



\begin{table}[t]
    \centering
    \renewcommand{\arraystretch}{1.}

    \caption{\textbf{Comparison on the multi/no object RES task.} Our models are highlighted with a gray background, and the top-performing method is shown in bold.  “Mask-Decoder-Free” indicates results without SAM-based post-processing.}
    \label{tab:gref}
    
    \definecolor{highlightgray}{gray}{0.93}
    \definecolor{citegray}{gray}{0.55}
    \newcommand{\citeinfo}[1]{{\color{citegray}\scriptsize #1}}

    \begin{adjustbox}{max width=\linewidth} 
    \setlength{\tabcolsep}{4pt}
    \begin{tabular}{@{}llrrrrrrr@{}}
        \toprule
        \multirow{2}{*}{\textbf{Method}} & \multirow{2}{*}{\textbf{LLM}} & \multicolumn{2}{c}{\textbf{Val Set}} & \multicolumn{2}{c}{\textbf{Test Set A}} & \multicolumn{2}{c}{\textbf{Test Set B}} & \multirow{2}{*}{\textbf{Avg.}} \\
        \cmidrule(lr){3-4} \cmidrule(lr){5-6} \cmidrule(lr){7-8}
        & & gIoU & cIoU & gIoU & cIoU & gIoU & cIoU & \\
        
        \midrule
        \multicolumn{9}{@{}l}{\quad \textit{Mask-Decoder-Free}} \\
        Text4Seg & InternLM2.5-7B & 70.0 & 66.1 & 69.4 & 70.9 & 63.1 & 64.1 & 67.3 \\
        Text4Seg & Vicuna-13B & 70.3 & 66.9 & 69.8 & 71.4 & 63.8 & 64.4 & 67.8 \\
        \rowcolor{highlightgray}
        \model & Qwen2-2B & 73.9 & 70.3 & 73.6 & 73.7 & 67.5 & 68.1 & 71.2 \\
        \rowcolor{highlightgray}
        \model & Qwen2-7B & \textbf{74.4} & \textbf{70.9} & \textbf{73.8} & \textbf{74.7} & \textbf{68.1} & \textbf{69.1} & \textbf{71.8}\\
        \addlinespace[0.3em]
        
        \multicolumn{9}{@{}l}{\quad \textit{With Mask Decoder}} \\
        LAVT & BERT & 58.4 & 57.6 & 65.9 & 65.3 & 55.8 & 55.0 & 59.7 \\
        LISA & Vicuna-7B & 61.6 & 61.8 & 66.3 & 68.5 & 58.8 & 60.6 & 62.9 \\
        ReLA & BERT & 63.6 & 62.4 & 70.0 & 69.3 & 61.0 & 59.9 & 64.4 \\
        LISA & Vicuna-13B & 63.5 & 63.0 & 68.2 & 69.7 & 61.8 & 62.2 & 64.7 \\
        GSVA & Vicuna-7B & 66.5 & 63.3 & 71.1 & 69.9 & 62.2 & 60.5 & 65.6 \\
        Text4Seg & InternLM2.5-7B & 74.4 & 69.1 & 75.1 & 73.8 & 67.3 & 66.6 & 71.1 \\
        Text4Seg & Vicuna-13B & 74.8 & 69.8 & 75.1 & 74.3 & 68.0 & 67.1 & 71.5 \\
        \rowcolor{highlightgray}
        \model & Qwen2-2B & 76.4 & 72.2 & 76.5 & 75.7 & 70.0 & 69.8 & 73.4 \\
        \rowcolor{highlightgray}
        \model & Qwen2-7B & \textbf{77.6} & \textbf{73.6} & \textbf{77.6} & \textbf{77.2} & \textbf{71.4} & \textbf{71.6} & \textbf{74.8}\\
        \bottomrule
    \end{tabular}
    \end{adjustbox}
\end{table}

\begin{table}[htbp]
    \centering
    \renewcommand{\arraystretch}{1}
    \caption{\textbf{Comparison on the reasoning segmentation task.}}
    \label{tab:reasonseg}
    
    \definecolor{highlightgray}{gray}{0.93}
    \definecolor{citegray}{gray}{0.55}
    \newcommand{\citeinfo}[1]{{\color{citegray}\scriptsize #1}}

    \begin{adjustbox}{max width=0.85\linewidth} 
    \begin{tabular}{@{}llrrrrr@{}}
        \toprule
        \multirow{2}{*}{\textbf{Method}} & \multirow{2}{*}{\textbf{LLM}} & \multicolumn{2}{c}{\textbf{Val}} & \multicolumn{2}{c}{\textbf{Test}} & \multirow{2}{*}{\textbf{Avg.}} \\
        \cmidrule(lr){3-4} \cmidrule(lr){5-6}
        & & gIoU & cIoU & gIoU & cIoU & \\
        \midrule
        OVSeg & Vicuna-7B & 28.5 & 18.6 & 26.1 & 20.8 & 23.5 \\
        LISA & Vicuna-7B & 53.6 & 52.3 & 48.7 & 48.8 & 50.9 \\
        SegLLM & Vicuna-7B & 57.2 & 54.3 & 52.4 & 48.4 & 53.1 \\
        Text4Seg & Qwen2-7B & 59.1 & 49.5 & 57.1 & 52.1 & 54.5 \\
        Seg-Zero & Qwen2.5-7B & 62.6 & 62.0 & 57.5 & 52.0 & 58.5 \\
        READ & Vicuna-7B & 59.8 & \textbf{67.6} & 58.5 & 58.6 & 61.1 \\
        \rowcolor{highlightgray}
        \model  & Qwen2-2B & \textbf{65.1} & 63.9 & \textbf{62.7} & \textbf{60.9} & \textbf{63.2} \\
        \bottomrule
    \end{tabular}
    \end{adjustbox}
\end{table}


\begin{table*}[htbp]
    \centering
    \renewcommand{\arraystretch}{0.97} 

    \caption{
       \textbf{Comparison of joint visual understanding and segmentation performance.} We include LLaVA-1.5-7B and Qwen2-VL-2B fine-tuned on the LLaVA-665k set (VQA) as reference models. The training data consists of segmentation data (Seg.), the VQA set, or a mixture (Mix). Our models are highlighted with a gray background; Performance of reference models is shown in {gray text}. The results show that \model{} maintains its general dialogue abilities after mixed training, while its segmentation performance is further improved.
    }
    \label{tab:vqa_res_seamless}
    
    \definecolor{highlightgray}{gray}{0.93}
    \definecolor{textgray}{gray}{0.65}

    \begin{adjustbox}{max width=0.7\linewidth} 
    \setlength{\tabcolsep}{5pt}
    \begin{tabular}{@{}ll rrr rr ccc@{}}
        \toprule
        \multirow{2}{*}{\textbf{Methods}} & \multirow{2}{*}{\textbf{Training Data}} & \multicolumn{5}{c}{\textbf{VQA}} & \multicolumn{3}{c}{\textbf{RES (val)}} \\
        \cmidrule(lr){3-7} \cmidrule(lr){8-10}
        & & MMMU & MMBench & MMStar & ScienceQA & TextVQA & RefC & RefC+ & RefCg \\
        \midrule
        \textcolor{textgray}{LLaVA-1.5-7B} & \textcolor{textgray}{VQA} & \textcolor{textgray}{35.7} & \textcolor{textgray}{66.5} & \textcolor{textgray}{33.1} & \textcolor{textgray}{68.4} & \textcolor{textgray}{55.0} & \textcolor{textgray}{n.a.} & \textcolor{textgray}{n.a.} & \textcolor{textgray}{n.a.} \\

        \textcolor{textgray}{Qwen2-VL-2B} & \textcolor{textgray}{VQA} & \textcolor{textgray}{\textbf{38.3}} & \textcolor{textgray}{66.5} & \textcolor{textgray}{42.1} & \textcolor{textgray}{70.2} & \textcolor{textgray}{\textbf{70.7}} & \textcolor{textgray}{n.a.} & \textcolor{textgray}{n.a.} & \textcolor{textgray}{n.a.} \\
        LISA{-7B} & Mix & 0 & 0 & 0 & 0 & 0 & 74.9 & 65.1 & 67.9 \\
        READ{-7B} & Mix & 1.1 & 0 & 14.4 & 23.2 & 22.6 & 78.1 & 68.4 & 70.1 \\
        Text4Seg{-7B} & Mix & 34.0 & 54.8 & 33.4 & 68.1 & 55.0 & 79.3 & 72.1 & 72.1 \\
        [3pt]
        \rowcolor{highlightgray}
         \textcolor{textgray}{\model{-2B}} & \textcolor{textgray}{Seg.} & \textcolor{textgray}{n.a.} & \textcolor{textgray}{n.a.} & \textcolor{textgray}{n.a.} & \textcolor{textgray}{n.a.} & \textcolor{textgray}{n.a.} & \textcolor{textgray}{81.9} & \textcolor{textgray}{77.1} & \textcolor{textgray}{78.5} \\
        \rowcolor{highlightgray}
        \model{-2B} & Mix & 37.8 & \textbf{68.7} & \textbf{42.4} & \textbf{72.6} & 69.7 & \textbf{82.2} & \textbf{77.3} & \textbf{79.0} \\
        \bottomrule
    \end{tabular}
    \end{adjustbox}
\end{table*}

\subsection{Visual Understanding}
\noindent\textbf{Setup.}
To verify that \model{} retains its general dialogue abilities, we evaluated its comprehensive performance on general vision-language tasks. We strictly followed the protocol established by Text4Seg~\citep{lan2024text4seg}. Specifically, our model was fine-tuned from a base model for two epochs on a composite dataset comprising 800k samples from the RefCOCO* suite and 665k vision-language instruction samples from LLaVA. For evaluation, we used the same referring segmentation benchmarks as Text4Seg to ensure a direct comparison. To assess visual understanding, we used a diverse range of VQA datasets, including MMMU~\cite{mmmu}, MMBench~\cite{liu2024mmbench}, MMStar~\cite{mmstar}, ScienceQA~\citep{lu2022learn} and TextVQA~\citep{singh2019towards}. Baseline methods are built upon the LLaVA; \model{} utilizes the Qwen backbone.

\noindent\textbf{Results.}
As shown in Table~\ref{tab:vqa_res_seamless}, training on the mixed dataset allowed \model{} to significantly outperform all baselines in both general-purpose visual understanding and segmentation simultaneously. Crucially, its visual understanding performance remained on par with the Qwen2-VL-2B baseline trained exclusively on VQA data. This demonstrates that our all-mask paradigm does not degrade the model's general dialogue abilities, effectively addressing the segmentation trilemma. Interestingly, this co-training strategy also boosted segmentation performance compared to training on segmentation-only data. This suggests that the all-mask prediction paradigm effectively allows higher-level understanding to enhance its segmentation ability, demonstrating a promising potential for further scaling.

\subsection{Ablation Study}

In this section, we conducted ablation studies to validate our architectural components and analyze the model's efficiency. We first ablated our key designs and evaluated the model's support for dynamic resolutions. We then benchmarked its inference speed against other paradigms to highlight high performance without sacrificing efficiency.

\noindent\textbf{Ablation of Core Components.}
As shown in~\cref{tab:my_awesome_table}, our two primary architectural designs are critical for performance. Removing either the visually-enhanced mask embedding (\(\mathbf{E}_{\text{mask}}\)) or the hybrid attention mechanism (\(\mathbf{A}_{\text{hyb}}\)) led to a distinct degradation in segmentation accuracy, confirming the contribution of each component.


\begin{table}[h]
\centering
\ra{0.9}
\caption{\textbf{Ablation study.} We report performance on the single-object RES task and the corresponding inference time.}
\label{tab:my_awesome_table}
\begin{adjustbox}{max width=0.7\linewidth} 
\begin{tabular}{@{}llrrr@{}}
\toprule
\textbf{Method} & \textbf{Input Size} & \textbf{cIoU} & \textbf{Time (s)} \\
\midrule
\model{-2B}                & $896 \times 896$       & 79.1   & 1.3          \\
\quad w/o $\mathbf{A}_{\text{hyb}}$  & $896\times896$          & 76.2   & 1.3      \\ 
\quad w/o $\mathbf{E}_{\text{mask}}$  & $896\times896$           & 73.3   & 1.3    \\
\quad w/o SAM         & $896\times896$  & 75.3   & 0.9      \\
\model{-2B}             & $726 \times 726$       & 78.6 & 1.1             \\
\model{-2B}             & $504 \times 504$       & 77.1 &  0.9            \\
\model{-7B}                 & $896 \times 896$       & 80.7 & 2.4              \\

\bottomrule
\end{tabular}
\end{adjustbox}
\end{table}

\noindent\textbf{Dynamic Resolution Support.}
The results in \cref{tab:my_awesome_table} also demonstrate our model's adaptability. Although trained at an 896 resolution, \model{} could process lower-resolution inputs, trading a marginal drop in performance for a substantial gain in inference speed. This versatility makes it suitable for diverse deployment scenarios.

\begin{figure}[t]
    \centering
    \includegraphics[width=0.8\linewidth]{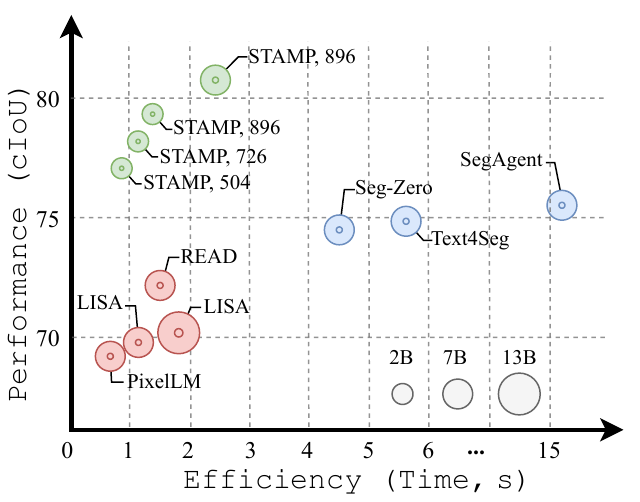}
    \caption{\textbf{Efficiency comparison.} Methods from the same paradigm are grouped by color. Numbers following the model names denote the input resolution. }
    \label{fig:speed}
    \vspace{-0.4em}
\end{figure}

\noindent\textbf{Efficiency Comparison.}
As shown in~\cref{fig:speed}, \model{} excels in both segmentation performance and speed. The inference speed is on par with the embedding approaches, clearly distinguishing it from slower next-token approaches.

\section{Conclusion}
In this work, we addressed the fundamental trilemma in MLLM-based segmentation, where models have been forced to trade off between general dialogue abilities, segmentation performance, and inference speed. We introduced all-mask prediction, a novel paradigm that resolves this conflict by decoupling autoregressive dialogue generation from non-autoregressive mask prediction. Our implementation, \model{}, is the first effective realization of this paradigm. Through extensive experiments, we demonstrated that \model{} not only establishes a new state of the art across multiple challenging benchmarks, including referring and reasoning segmentation, but also achieves this with inference speeds comparable to the efficient prior paradigms, all while fully preserving its general dialogue abilities. By successfully resolving the segmentation trilemma, our work removes the barrier to developing more practical, efficient, and truly unified visual models.

{
\small
\bibliographystyle{ieeenat_fullname}
\bibliography{main}
}

\clearpage
\setcounter{section}{0}
\renewcommand{\thesection}{S\arabic{section}}

\setcounter{figure}{0}  
\renewcommand{\thefigure}{S\arabic{figure}}

\setcounter{table}{0}   
\renewcommand{\thetable}{S\arabic{table}}

\setcounter{equation}{0}
\renewcommand{\theequation}{S\arabic{equation}}

\setcounter{page}{1}

\renewcommand{\theHsection}{S\arabic{section}}
\renewcommand{\theHfigure}{S\arabic{figure}}
\renewcommand{\theHtable}{S\arabic{table}}

\maketitlesupplementary

In the supplementary materials, we report:
\begin{itemize}
    \item \textbf{Implementation Details.} We provide comprehensive implementation details, including the training parameters and datasets for each experimental stage, the methodology for constructing instruction-following data, and a comparison of datasets used by our method versus baseline models (\cref{sec:imp}).

    \item \textbf{Additional Experiments.} We report detailed analyses of our model's efficiency, including inference time, GPU memory consumption (\cref{sec:exp}).

    \item \textbf{More Qualitative Showcases.} We present detailed qualitative showcases of \model, covering standard referring segmentation(in the main text), reasoning segmentation, visual question answering and advanced capabilities such as muti-round dialogue, multi-round segmentation and unified dialogue-segmentation examples (\cref{sec:show}).

    \item \textbf{Code and Documentation.} We provide the source code for training and execution, along with a user guide and setup instructions (\cref{sec:files}).
\end{itemize}

\section{Implementation Details}
\label{sec:imp}

In this section, we detail our training strategies, data construction methodology, and the experimental setup for a fair comparison against baseline methods.

\subsection{Training Strategy}

The specific training hyperparameters for the experiments in each section are detailed in \cref{tab:unified_sft_stages}, aligning with standard practices. Our experiments utilize the Qwen2-VL-2B and Qwen2-VL-7B models~\citep{bai2025qwen2}. We maintain consistent hyperparameters across experiments wherever possible to isolate the performance gains from our proposed methods, rather than from extensive hyperparameter tuning.

\noindent\textbf{LoRA Training.} We employ LoRA~\citep{hu2022lora} for training our 7B model. This decision aligns with the training strategies of baseline methods such as LISA~\citep{lai2024lisa} and Text4Seg~\citep{lan2024text4seg}, ensuring a direct and fair comparison. In contrast, our 2B model is fully fine-tuned, as its smaller size renders the efficiency benefits of LoRA~\citep{hu2022lora} less significant.

\subsection{Training Dataset}
For the training data in each sub-experiment, we strictly adhere to the settings of our baselines, Text4Seg~\citep{lan2024text4seg} and LISA~\citep{lai2024lisa}, conducting experiments on the same minimal datasets (detailed in \cref{tab:cpt_vs_sft_comparison}). This approach is intentional: we aim to demonstrate that our performance gains stem from our paradigm innovation, rather than from simply using more data. Consequently, our current training is focused on single-object grounding and segmentation tasks. While many contemporary works train on larger, mixed datasets, we are confident that \model can be scaled to these data sources in the future, presenting a clear path to even greater performance leads.

\subsection{Instruction Construction}
The instruction used for training \model is included in Prompt S1 and S2.

\begin{lstlisting}[caption={Instructions for RES task.}]
Question:
o "Please segment only the [class_name] in the image.",
o "Can you segment the [class_name] in the image?",
o "Where is the [class_name] in this picture? Please respond with segmentation mask.",
o "Where is [class_name] in this image? Please output segmentation mask.",
o "Could you provide the segmentation mask for [class_name] in this image?",
o "Please segment the image and highlight [class_name]."

Answer:
* "Sure, here is the segmentation mask for [class_name] <|seg|>:",
* "Here is the segmentation mask focusing on the [class_name] <|seg|>:",
* "Here is the segmentation mask highlighting the [class_name] <|seg|>:",
* "The segmentation map for [class_name] <|seg|> is:",
* "The segmentation mask for [class_name] <|seg|> is shown below:",
* "Sure, Here's the segmentation mask for [class_name] <|seg|>:",
* "Sure, the segmented output for [class_name] <|seg|> is:",
* "Certainly, the segmentation map for [class_name] <|seg|> is:",
* "Certainly, here is the segmentation mask for [class_name]:",
* "The segmentation mask for [class_name] <|seg|> is shown below:"
\end{lstlisting}

\begin{lstlisting}[caption={Instructions for Reasoning Segmentation task.}]
% --- 1. Segmentation Task ---
Question:
    o "Can you segment the [class_name] in this image?" 
    o "Please segment the [class_name] in this image."
    o "What is [class_name] in this image? Please output segmentation mask."
    o "{sent} Please respond with segmentation mask."

Answer:
    * "It is <|seg|>."
    * "Sure, <|seg|>."
    * "Sure, the segmentation result is <|seg|>."
    * "<|seg|>."

% --- 2. Explanation Task ---
Question:
o "Explain [class_name] in this image."
o "Could you describe the [class_name] in the image?"
o "Tell me more about the [class_name]."

Answer:
* "[reasoning]" 
\end{lstlisting}

\begin{table*}[t]
\centering
\caption{\textbf{Comprehensive hyperparameter settings for all SFT stages.} 
We detail the training configurations for our main experiments: Referring Expression Segmentation (\S4.2), Reasoning Segmentation (\S4.3), and Visual Understanding (\S4.4). RefCOCO* means RefCOCO~\citep{kazemzadeh2014referitgame}, RefCOCO+~\citep{kazemzadeh2014referitgame}, RefCOCOg~\citep{mao2016generation} and RefCLEF~\citep{kazemzadeh2014referitgame}.}
\label{tab:unified_sft_stages}
\adjustbox{max width=1\textwidth}{
\renewcommand{\arraystretch}{1.1} 
\setlength{\tabcolsep}{5pt}     
\begin{tabular}{@{}llcccc@{}}
\toprule
\multirow{2}{*}{\textbf{Category}} & \multirow{2}{*}{\textbf{Hyperparameter}} & \multicolumn{2}{c}{\textbf{Referring Seg.}} & {\textbf{Reasoning Seg.}} & \textbf{Dialogue \& Seg.} \\
\cmidrule(lr){3-4} \cmidrule(lr){5-5} \cmidrule(lr){6-6}
& & \textbf{Qwen2{-2B}} & \textbf{Qwen2{-7B}} & \textbf{Qwen2{-2B}} & \textbf{Qwen2{-2B}} \\
\midrule
\multirow{5}{*}{Training Details}
 & Section \S & \S4.2 & \S4.2 & \S4.3 & \S4.4 \\[3pt]
 & Datasets & \makecell{RefCOCO*, \\ gRefCOCO~\citep{liu2023gres} (stage 2)} & \makecell{RefCOCO*, \\ gRefCOCO~\citep{liu2023gres} (stage 2)} & \makecell{RefCOCO*,\\ ReasonSeg~\citep{lai2024lisa}} & \makecell{RefCOCO*, \\ LLaVA-665k~\citep{llava} } \\[3pt]
 & Epochs & 3 + 2 & 3 + 2 & 1 & 2 \\
 & Batch Size & 96 & 96 & 96 & 96 \\
 & ZeRO Stage & 2 & 2 & 2 & 2 \\
\midrule
\multirow{6}{*}{Optimizer (AdamW)}
 & Learning Rate & \num{3e-5} & \num{2e-5} & \num{3e-5} & \num{3e-5} \\
 & Weight Decay & 0 & 0 & 0 & 0 \\
 & Betas ($\beta_1, \beta_2$) & (0.9, 0.999) & (0.9, 0.999) & (0.9, 0.999) & (0.9, 0.999) \\
 & Grad Norm Clip & 1.0 & 1.0 & 1.0 & 1.0 \\
 & Scheduler & Linear & Linear & Linear & Linear \\
 & Warmup Ratio & 0.03 & 0.03 & 0.03 & 0.03 \\
\midrule
\multirow{5}{*}{LoRA Configuration}
 & Method & \multirow{5}{*}{\makecell{Full fine-tuning \\ (LoRA not applied)}} & LoRA & \multirow{5}{*}{\makecell{Full fine-tuning \\ (LoRA not applied)}} & \multirow{5}{*}{\makecell{Full fine-tuning \\ (LoRA not applied)}} \\
 & Rank &  & 64 &  &  \\ 
 & Alpha ($\alpha$) &  & 128 &  &  \\
 & Dropout &  & 0.05 &  &  \\
 & Target Modules &  & All &  &  \\
\midrule
Model Parameters
 & Total Parameters & 2B & 7B & {2B} & 2B\\
\bottomrule
\end{tabular}
}
\end{table*}

\begin{table*}[t]
\centering
\caption{\textbf{Training data comparison across CPT and SFT stages.} 
This table contrasts the datasets used during the Continued Pre-Training (CPT) stage and the Supervised Fine-Tuning (SFT) stage for Referring Expression Segmentation. 
\dag\ Some methods incorporate RefCLEF during CPT, while others (like Ours, Text4Seg, \model) use it in the SFT stage.}
\label{tab:cpt_vs_sft_comparison}
\adjustbox{max width=0.85\textwidth}{
\renewcommand{\arraystretch}{1.25} 
\setlength{\tabcolsep}{6pt}      
\begin{tabular}{@{}l | p{0.5\textwidth} | p{0.4\textwidth} @{}}
\toprule
\textbf{Method} & \textbf{Continued Pre-Training (CPT) Datasets} & \textbf{Referring Seg. SFT Datasets} \\
\midrule
LISA~\citep{lai2024lisa} & ADE20K~\citep{zhou2019semantic}, COCO-Stuff~\citep{caesar2018coco}, PACO-LVIS~\citep{ramanathan2023paco}, PartImageNet~\citep{he2022partimagenet}, PASCAL-Part~\citep{chen2014detect}, RefCLEF\dag~\citep{kazemzadeh2014referitgame}, RefCOCO~\citep{kazemzadeh2014referitgame}, RefCOCO+~\citep{kazemzadeh2014referitgame}, RefCOCOg~\citep{mao2016generation}, LLaVA-665k~\citep{llava} & RefCOCO~\citep{kazemzadeh2014referitgame}, RefCOCO+~\citep{kazemzadeh2014referitgame}, RefCOCOg~\citep{mao2016generation} \\
\cmidrule(lr){1-3}
PixelLM~\citep{ren2024pixellm} & ADE20K~\citep{zhou2019semantic}, COCO-Stuff~\citep{caesar2018coco}, PACO-LVIS~\citep{ramanathan2023paco}, RefCLEF\dag~\citep{kazemzadeh2014referitgame}, RefCOCO~\citep{kazemzadeh2014referitgame}, RefCOCO+~\citep{kazemzadeh2014referitgame}, RefCOCOg~\citep{mao2016generation}, LLAVA-150k~\citep{llava}, MUSE~\citep{ren2024pixellm} & \textbf{None} \\
\cmidrule(lr){1-3}
GSVA~\citep{xia2024gsva} & ADE20K~\citep{zhou2019semantic}, COCO-Stuff~\citep{caesar2018coco}, PACO-LVIS~\citep{ramanathan2023paco}, Mapillary Vistas~\citep{neuhold2017mapillary}, PASCAL-Part~\citep{chen2014detect}, RefCLEF\dag~\citep{kazemzadeh2014referitgame}, RefCOCO~\citep{kazemzadeh2014referitgame}, RefCOCO+~\citep{kazemzadeh2014referitgame}, RefCOCOg~\citep{mao2016generation}, gRefCOCO~\citep{liu2023gres}, LLaVA-Instruct-150K~\cite{llava}, ReasonSeg~\citep{lai2024lisa} & RefCOCO~\citep{kazemzadeh2014referitgame}, RefCOCO+~\citep{kazemzadeh2014referitgame}, RefCOCOg~\citep{mao2016generation} \\
\cmidrule(lr){1-3}
READ~\citep{read} & ADE20K~\citep{zhou2019semantic}, COCO-Stuff~\citep{caesar2018coco}, PACO-LVIS~\citep{ramanathan2023paco}, PASCAL-Part~\citep{chen2014detect}, RefCLEF\dag~\citep{kazemzadeh2014referitgame}, RefCOCO~\citep{kazemzadeh2014referitgame}, RefCOCO+~\citep{kazemzadeh2014referitgame}, RefCOCOg~\citep{mao2016generation}, LLaVA-Instruct-150K~\cite{llava} & R-RefCOCO~\citep{wu2024toward}, FP-RefCOCO+~\citep{wu2024see}, FP-RefCOCOg~\citep{wu2024see} \\
\cmidrule(lr){1-3}
Seg-Zero~\citep{liu2025seg} & \textbf{None} & RefCOCOg~\citep{wu2024see} \\
\cmidrule(lr){1-3}
SegLLM~\citep{wang2025segllm} & RefCOCO~\citep{kazemzadeh2014referitgame}, Visual Genome~\citep{krishna2017visual}, PACO-LVIS~\citep{ramanathan2023paco}, LVIS~\citep{gupta2019lvis}, Pascal Panoptic Part~\citep{de2021part}, ADE20K~\citep{zhou2019semantic}, COCO-Stuff~\citep{caesar2018coco}, Attributes-COCO~\citep{lin2014microsoft}, ReasonSeg~\citep{lai2024lisa}, MRSeg (hard)~\citep{wang2025segllm} & \textbf{None} \\
\cmidrule(lr){1-3}
SegAgent~\citep{zhu2025segagent} & \textbf{None} & DIS5K~\citep{qin2022highly}, ThinObject5K~\citep{liew2021deep} \\
\cmidrule(lr){1-3}
Text4Seg~\citep{lan2024text4seg} & \textbf{None} & RefCOCO~\citep{kazemzadeh2014referitgame}, RefCOCO+~\citep{kazemzadeh2014referitgame}, RefCOCOg~\citep{mao2016generation}, RefCLEF\dag~\citep{kazemzadeh2014referitgame} \\
\cmidrule(lr){1-3}
\rowcolor{gray!15}
\model & \textbf{None} & RefCOCO~\citep{kazemzadeh2014referitgame}, RefCOCO+~\citep{kazemzadeh2014referitgame}, RefCOCOg~\citep{mao2016generation}, RefCLEF\dag~\citep{kazemzadeh2014referitgame} \\
\bottomrule
\end{tabular}
}
\end{table*}









\section{Additional Experiments.}
\label{sec:exp}
In this section, we list the specific design and detailed test results for evaluating \model's efficiency.

\subsection{Implementation Setting.}
\begin{figure}[t]
    \centering
    \includegraphics[width=\columnwidth]{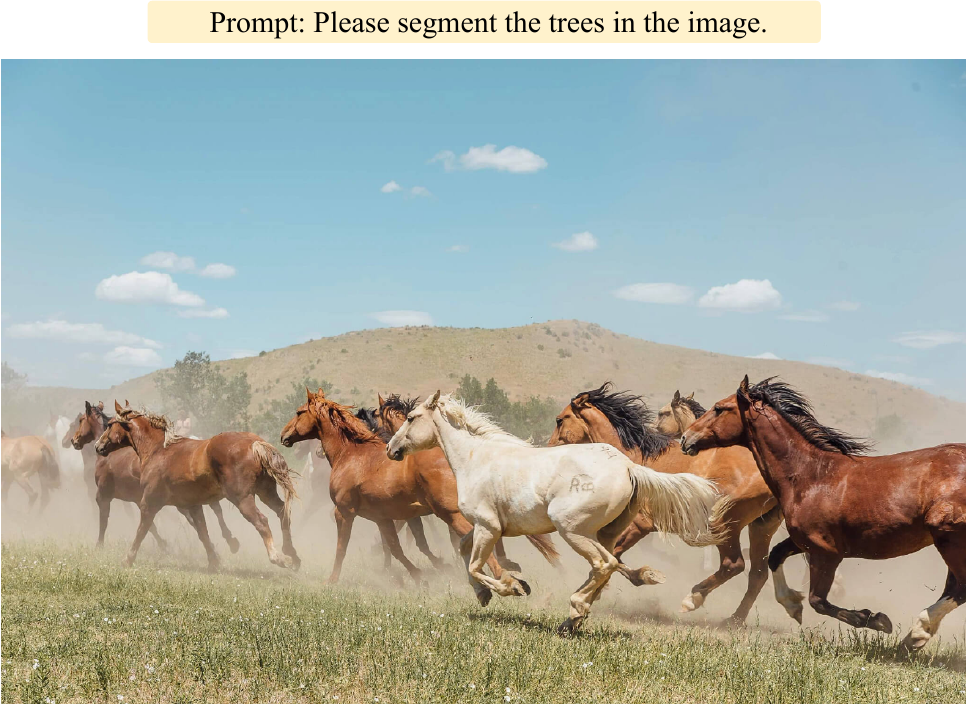}
    \caption{\textbf{Inference time comparison example}. This example illustrates a representative test case used to measure the inference latency across all compared methods. The consistent input allows for a fair comparison of the computational efficiency between the models discussed in Section 4.5.}
    \label{fig:inference_time_example}
\end{figure}

For the inference latency discussed in \S4.5, all methods were rigorously evaluated on a single A800 80GB GPU. Crucially, tests were conducted in isolation, ensuring only one program ran on the card during each measurement. Furthermore, all methods employed the identical test case, which is illustrated in ~\cref{fig:inference_time_example}.

\subsection{Results.}
As shown in \cref{tab:ablation_and_efficiency}, \model excels in both accuracy and speed. \model{-7B} achieves the highest cIoU of 80.7, significantly outperforming SegAgent{-7B}(75.4)~\citep{zhu2025segagent} and READ{-7B}(71.9)~\citep{read}. Despite high input resolution ($896 \times 896$), \model remains remarkably fast: \model{-2B} variants (0.9s--1.3s) are comparable to embedding-based methods (e.g., LISA, PixelLM~\citep{ren2024pixellm}) and far faster than next-token approaches like SegAgent (15.9s). Ablation studies further confirm this efficiency; even without the mask decoder, \model{-2B} maintains competitive performance (75.3 cIoU) with minimal VRAM usage (10.8 GB), demonstrating superior resource efficiency.

\begin{table*}[t]
\centering
\caption{\textbf{Ablation and efficiency study on the single-object RES task.} 
We compare performance on the RES task (cIoU), inference time, and decoder dependency across paradigms. Methods are grouped by their core segmentation paradigm as defined in Table 1. 
All inference metrics (Time and VRAM) are benchmarked on an A800 GPU.}
\label{tab:ablation_and_efficiency}
\begin{adjustbox}{max width=0.75\linewidth}
\renewcommand{\arraystretch}{1.2}
\setlength{\tabcolsep}{6pt}
\begin{tabular}{@{}lrrcrc@{}}
\toprule
\textbf{Method} & 
\textbf{Input Size} & 
\textbf{cIoU}& 
\textbf{Time (s)} & 
\textbf{Decoder Status} & 
\textbf{Infer VRAM (GB)}\\
\midrule
\multicolumn{6}{@{}l}{\textit{Paradigm 1: Embedding Prediction}} \\
LISA{-7B} \small{(CVPR'24)}~\citep{lai2024lisa}         & $336 \times 336$ & 69.9 & 1.1 & Required SAM & 15.5 \\
LISA-13B \small{(CVPR'24)}~\citep{lai2024lisa}        & $336 \times 336$ & 70.1 & 1.8 & Required SAM & 27.6 \\
READ{-7B} \small{(CVPR'25)}~\citep{read}         & $336 \times 336$ & 71.9 & 1.5 & Required SAM & 15.7 \\
PixelLM{-7B} \small{(CVPR'24)}~\citep{ren2024pixellm}      & $448 \times 448$ & 69.2 & 0.7 & Required Custom Dec. & 14.1 \\
\midrule

\multicolumn{6}{@{}l}{\textit{Paradigm 2: Next-token Prediction}} \\
Seg-Zero{-7B} \small{(arXiv'25)}~\citep{liu2025seg}     & $840 \times 840$ & 74.1    & 4.5 & Required SAM2   & 18.0 \\
SegAgent{-7B} \small{(CVPR'25)}~\citep{zhu2025segagent}     & $448 \times 448$ & 75.4 & 15.9 & Required SAM    & 20.7 \\
Text4Seg{-7B} \small{(ICLR'25)}~\citep{lan2024text4seg}     & $336 \times 336$ & 74.9 & 4.0 & Optional         & 20.4 \\
\quad \textit{W/O Mask Decoder} 
                & $336 \times 336$ & 70.9 & 3.9 & \emph{SAM}              & 20.4 \\
\midrule

\multicolumn{6}{@{}l}{\textit{Paradigm 3: All-mask Prediction (Ours)}} \\
\rowcolor{gray!15}
\model{-2B}        & $504 \times 504$ & 77.1 & 0.9 & Optional & 10.8 \\
\rowcolor{gray!15}
\model{-2B}        & $726 \times 726$ & 78.6 & 1.1 & Optional & 10.8 \\
\rowcolor{gray!15}
\model{-2B}        & $896 \times 896$ & 79.1 & 1.3 & Optional & 10.8 \\
\rowcolor{gray!15}
\quad \textit{W/O Mask Decoder} 
                & $896 \times 896$ & 75.3 & 0.9 & \emph{SAM}  & 10.8 \\
\rowcolor{gray!15}
\model{-7B}        & $896 \times 896$ & 80.7 & 2.4 & Optional & 25.6 \\
\rowcolor{gray!15}
\quad \textit{W/O Mask Decoder} 
                & $896 \times 896$ & 76.2 & 2.1 & \emph{SAM}  & 25.6 \\
\bottomrule
\end{tabular}
\end{adjustbox}
\end{table*}

\section{More Qualitative Showcases.}
To comprehensively validate the capabilities of our \model model, we designed a suite of test cases targeting its performance across multiple areas, including reasoning segmentation, VQA (Visual Question Answering), multi-round conversation, multi-round segmentation, and dialogue-segmentation. For all subsequent cases, the \model model used is the \model{-2B} variant. This model is configured with the SAM decoder and processes inputs at an $896 \times 896$ resolution.
\label{sec:show}

\subsection{Reasoning Segmentation}
\begin{figure*}[t]
    \centering
    \includegraphics[width=0.9\linewidth]{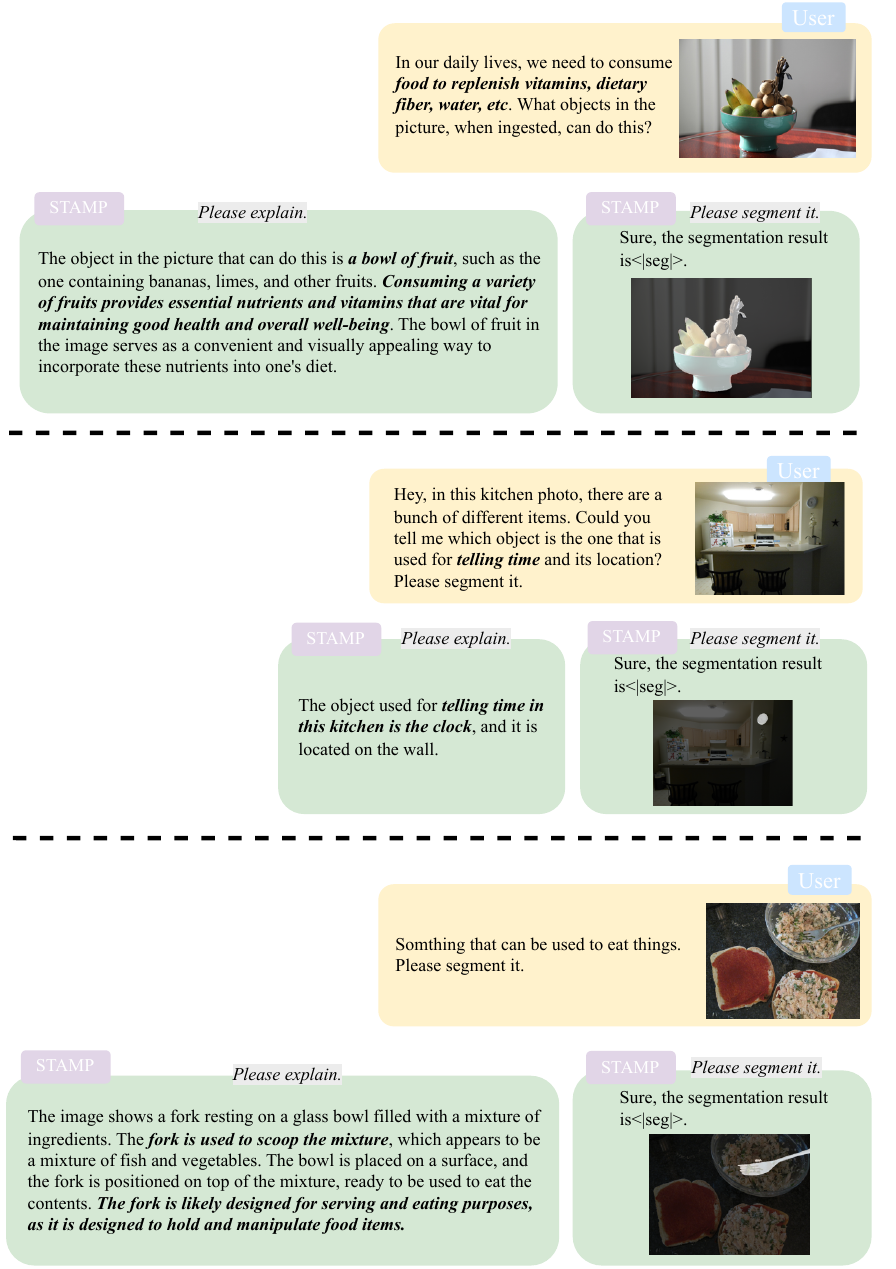}
    \caption{\textbf{Reasoning segmentation cases.} The left column presents the model's detailed explanations of inferred object properties, such as the nutritional role of fruit, the time-keeping function of a clock, and the utility of a fork. Correspondingly, the right column displays the successful segmentation results derived from this understanding."}
    \label{fig:reasonseg}
\end{figure*}

Reasoning Segmentation requires transcending simple recognition to integrate functional and abstract reasoning. As shown in \cref{fig:reasonseg}, \model excels in this task by explicitly decoupling reasoning from segmentation. It first articulates inferred properties (e.g., nutritional role) before executing precise segmentation. This two-step process confirms \model's ability to ground high-level semantic understanding directly onto pixel space, offering a distinct advantage over models that treat reasoning merely as a textual side effect.
\subsection{Visual Question Answering}
\begin{figure*}[t]
    \centering
    \includegraphics[width=0.77\linewidth]{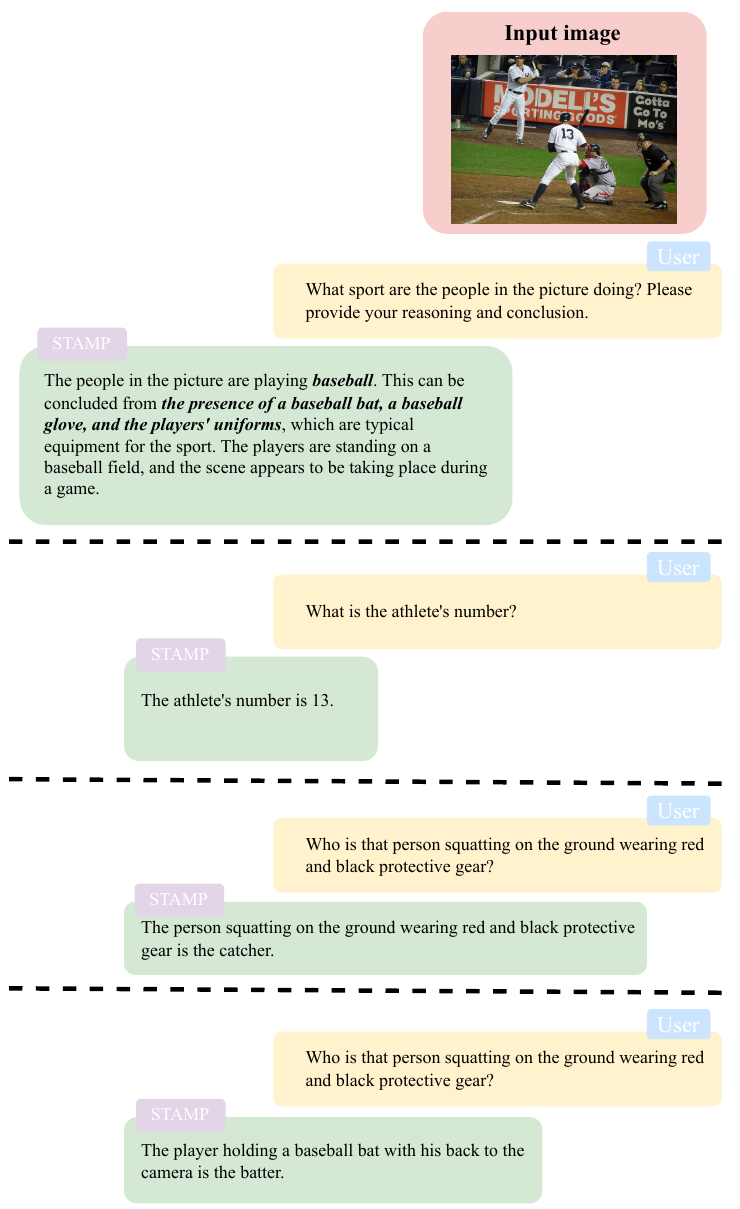}
    \caption{\textbf{Visual question answering (VQA) cases.} This example showcases \model's performance across various single-turn VQA scenarios. Each exchange, separated by a dashed line, is treated as an independent question-answering task. The model demonstrates comprehensive visual understanding, covering: 1) complex scene reasoning (identifying the sport), 2) precise detail extraction (athlete's number), and 3) specific referential grounding (identifying the catcher and batter).}
    \label{fig:vqa}
\end{figure*}

Crucially, \model integrates vision-language capabilities without sacrificing the core language proficiency of the underlying MLLM. This contrasts with methods like LISA that often compromise language performance for segmentation. As shown in \cref{fig:vqa}, \model remains robust across diverse VQA tasks, ranging from abstract reasoning to fine-grained grounding. This demonstrates that \model retains superior language mastery, serving as a versatile multimodal assistant beyond its primary segmentation role.
\subsection{Muti-round Dialogue}
\begin{figure*}[t]
    \centering
    \includegraphics[width=0.9\textwidth]{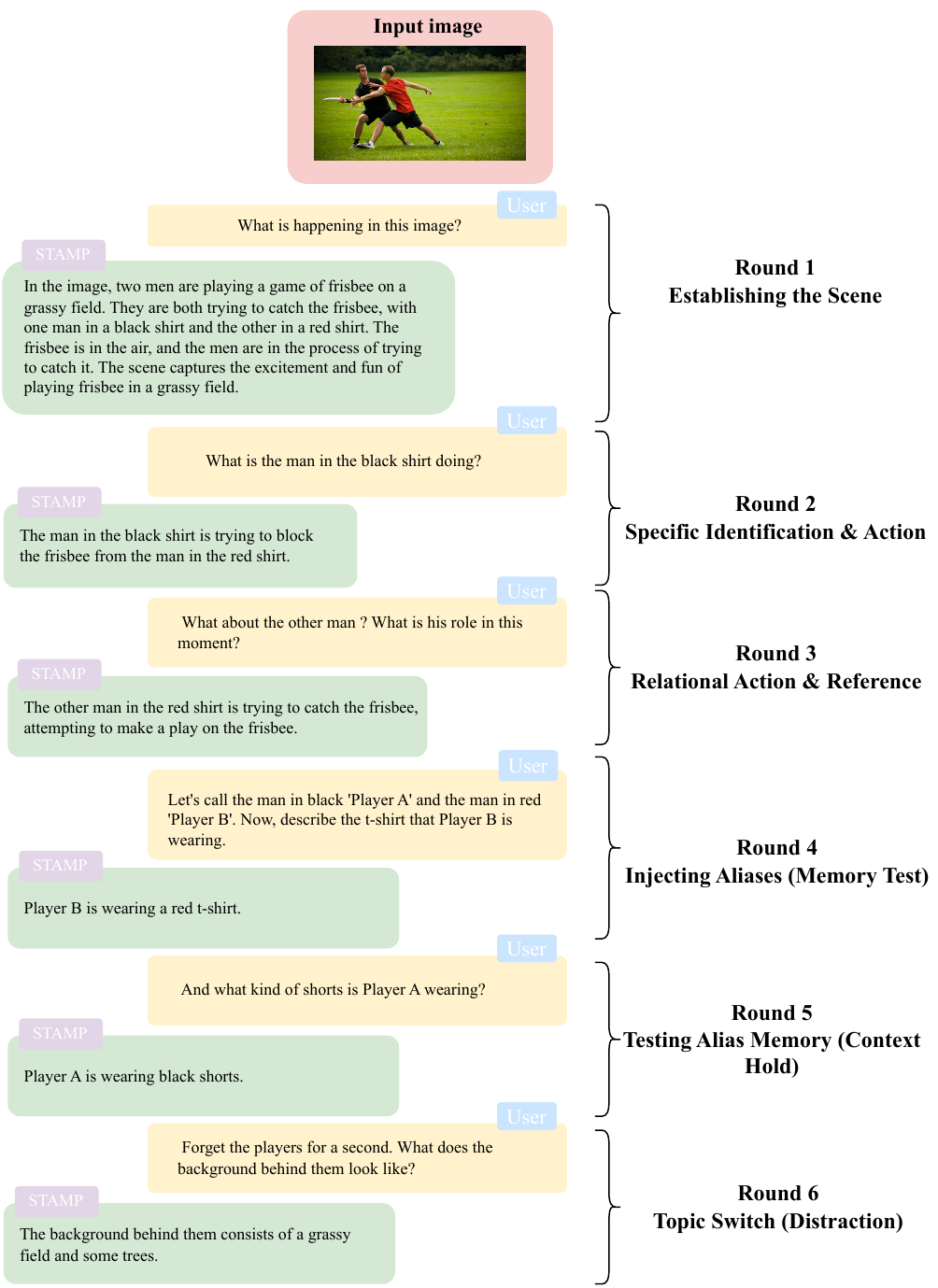}
    \caption{\textbf{Multi-round conversational capabilities.} This figure illustrates \model's performance in a complex, multi-turn conversation about a single image. The interaction is structured to test the model's ability to maintain context, including: establishing the scene (Round 1), specific identification (Round 2), relational referencing (Round 3), memory injection/testing via aliases (Rounds 4-5), and topic switching with distraction (Round 6). Successful completion of these rounds validates \model's robust contextual memory and dialogue coherence.}
    \label{fig:muti-round-conv}
\end{figure*}

The multi-round dialogue scenario rigorously tests \model's contextual retention and linguistic integrity. As illustrated in ~\cref{fig:muti-round-conv}, the model's seamless performance across a six-round exchange confirms its capability as a unified conversational agent. Specifically, \model demonstrates zero-shot immunity to common MLLM failures: it successfully grounds complex temporary aliases and maintains resilience against conversational distractions. This confirms that \model preserves the critical memory and reasoning capabilities essential for conversational AI, despite the integration of segmentation.

\subsection{Muti-round Segmentation}
The Multi-round Dialogue scenario evaluates \model's contextual retention and linguistic integrity. As shown in \cref{fig:muti-round-conv}, its seamless performance across a six-round exchange confirms its capability as a unified conversational agent. Specifically, \model demonstrates zero-shot robustness against common failures, successfully grounding temporary aliases and resisting conversational distractions. This sustained coherence validates that integrating segmentation preserves the critical memory and reasoning foundations required for advanced conversational AI.
\begin{figure*}[t]
    \centering
    \includegraphics[width=0.95\textwidth]{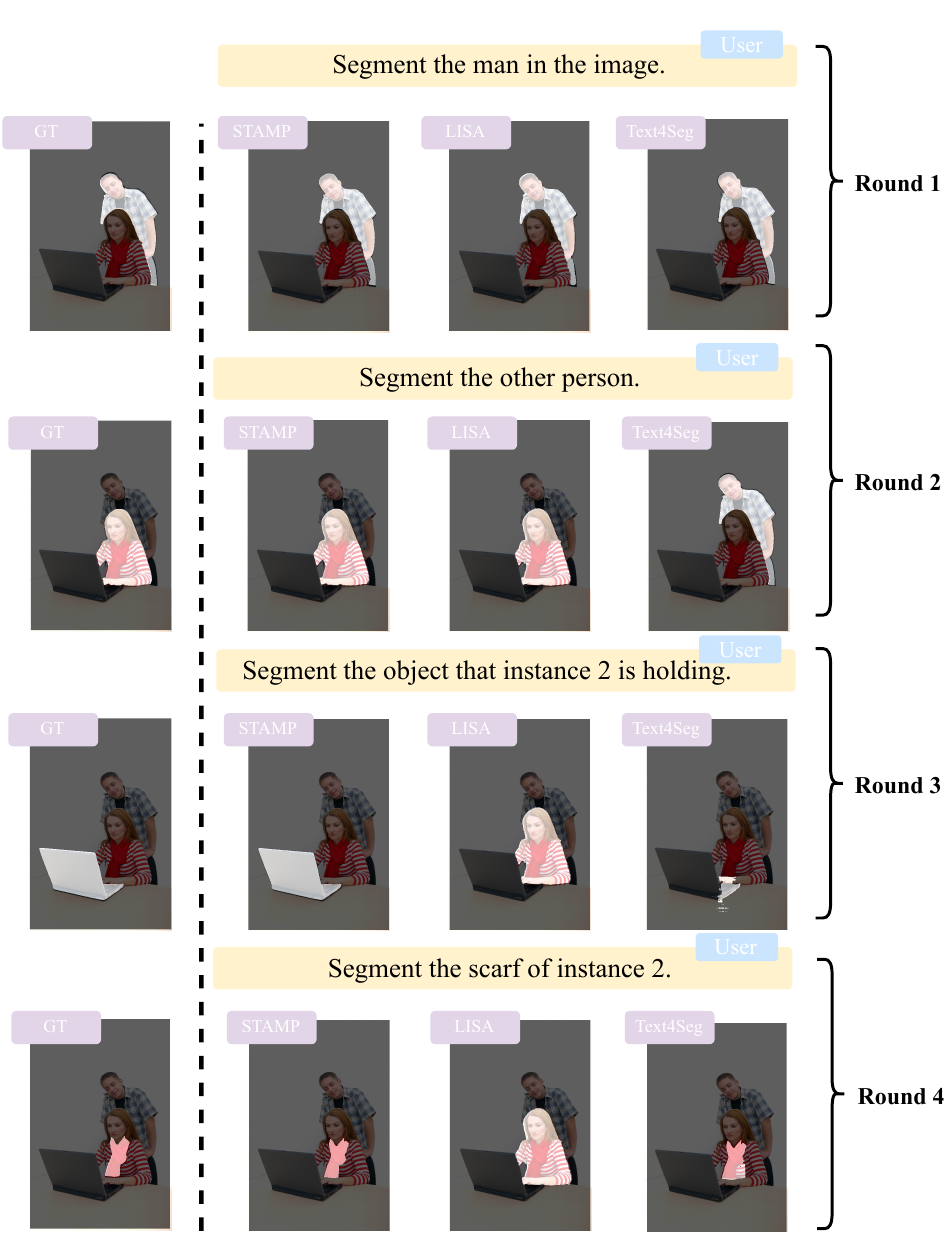}
    \caption{\textbf{Multi-round segmentation cases.} This figure illustrates \model's performance in multi-turn interactive segmentation, compared against LISA and Text4Seg. The sequential rounds (Round 1 to Round 4) require the model to maintain segmentation context and handle complex referential queries. \model consistently delivers accurate segmentation even as the queries become more nuanced (e.g., referencing "the other person" in Round 2, segmenting an object held by a previously identified instance in Round 3, and fine-grained attribute segmentation like "scarf of instance 2" in Round 4), showcasing its superior contextual understanding compared to other methods.}
    \label{fig:muti-round-segmentation}
\end{figure*}

\subsection{Dialogue Segmentation Interaction}
\begin{figure*}[t]
    \centering
    \includegraphics[width=\textwidth]{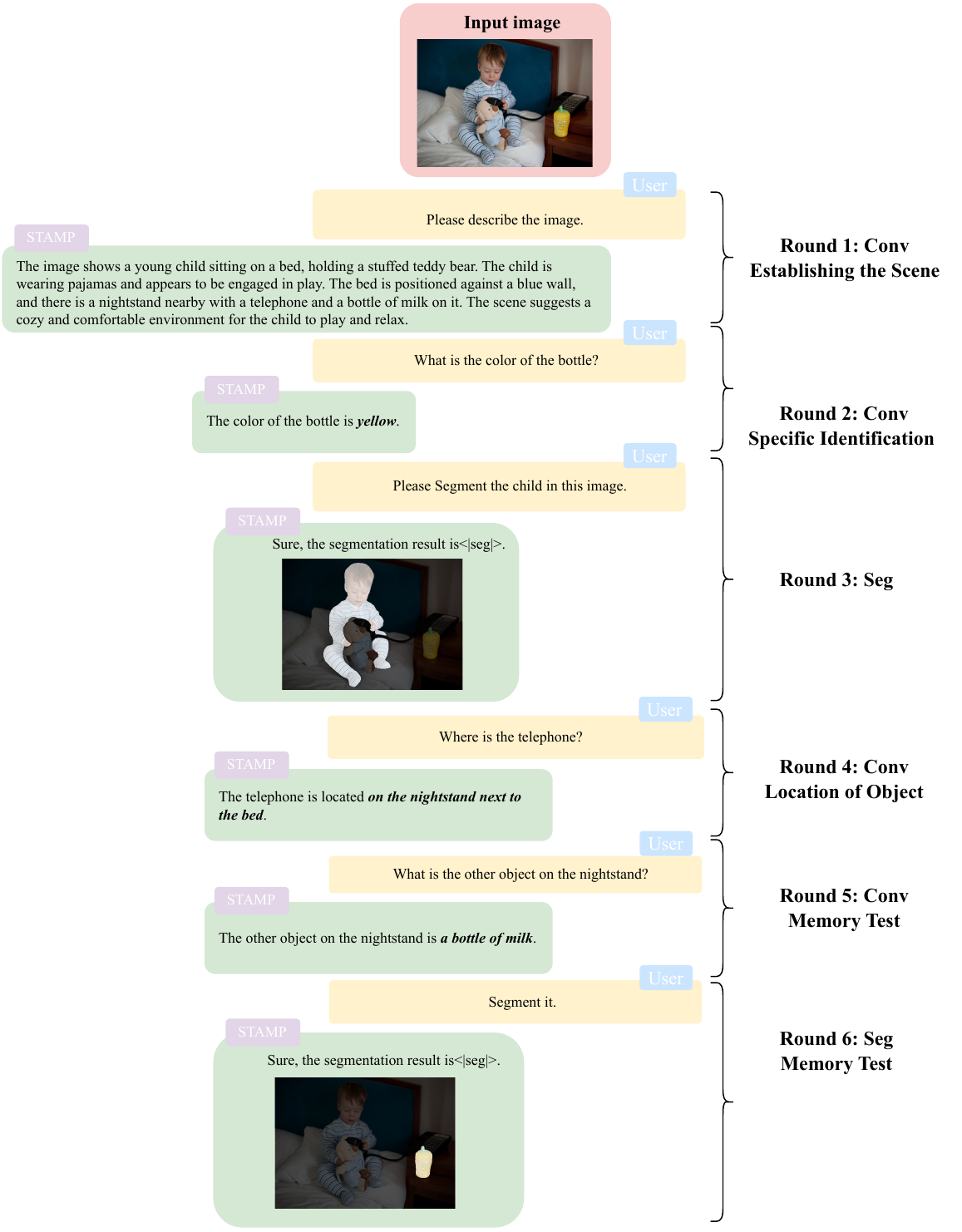}
    \caption{\textbf{Seamless interleaving of segmentation and dialogue.} It shows \model's unique ability to flexibly interleave Conv and Seg tasks in one dialogue thread. This structure demonstrates context-dependent segmentation, culminating in memory-based segmentation (Round 6). The successful segmentation in Round 6 confirms \model accurately grounds an object (bottle of milk) into pixel space using only the dialogue history (Rounds 4-5), without requiring a new descriptive prompt.}
    \label{fig:seg-conv}
\end{figure*}

The Dialogue Segmentation Interaction scenario validates \model as a unified multimodal assistant. As shown in \cref{fig:seg-conv}, the model seamlessly transitions between linguistic responses and pixel-level masks within the same flow. A key highlight is the memory-based segmentation in Round 6, where \model grounds an object (the milk bottle) based solely on context established in Rounds 4-5. This confirms \model utilizes dialogue history as a dynamic context for visual tasks.

\section{Code and Documentation.}
We provide a detailed \emph{README.md} as the primary reference for the codebase, covering essential aspects including environment setup, detailed usage protocols for training and evaluation, and an explanation of the modular project structure. Due to upload size constraints, raw image datasets are not included in this submission; however, the documentation strictly defines the expected data hierarchy. To facilitate reproducibility, we also include specific scripts designed to generate all required annotations from standard data sources.
\label{sec:files}




\end{document}